\documentclass{article}

\PassOptionsToPackage{numbers, sort&compress}{natbib}
\usepackage[preprint]{neurips_2026}

\usepackage[utf8]{inputenc}
\usepackage[T1]{fontenc}
\usepackage{url}
\usepackage{booktabs}
\usepackage{array}
\usepackage{amsfonts}
\usepackage{amsmath}
\usepackage{nicefrac}
\usepackage{microtype}
\usepackage{xcolor}
\usepackage{graphicx}
\usepackage{multirow}
\usepackage{pgfplots}
\pgfplotsset{compat=1.18}
\usetikzlibrary{arrows.meta, positioning}
\usepackage[hidelinks]{hyperref}

\title{Trading Human Curation for Synthetic Augmentation in RLVR}

\author{%
  Akshansh \\ Pareto AI \\ \texttt{akshansh@pareto.ai}
  \And
  Leonardo Rosa Rodrigues \\ Pareto AI \\ \texttt{leonardo@pareto.ai}
  \And
  Michael Korostelev \\ Pareto AI \\ \texttt{michael.korostelev@pareto.ai}
  \And
  Youssef Hassan \\ Pareto AI \\ \texttt{youssef@pareto.ai}
  \And
  Mark E.\ Whiting \\ Pareto AI \\ \texttt{mark@pareto.ai}
}

\begin{document}

\maketitle

\begin{abstract}
The supply of high-quality training tasks is one of the central bottlenecks for reinforcement learning from verifiable rewards (RLVR) on agentic language models. Each task requires a sandboxed setup, a prompt, and a hand-authored reward function, and only tasks that pass a quality bar produce useful training signal. Hand-curation at this quality bar does not scale in data-curation cost to the task counts effective RL training requires, and the substitution rate between automatically generated task variants and human-authored ones is not yet established. We investigate using pre-specified, gate-filtered augmentations of a small hand-authored base as a substitute for additional human curation during RLVR. We formalize the cost-adjusted trade rate $\rho_{\text{cost}}$ between augmented and human-authored tasks, measure it through a controlled ablation across training corpora with varying augmentation share, and characterize the end-to-end economics of the augmentation pipeline. Substituting augmented content for additional human-authored tasks retains aggregate held-out generalization on a ten-benchmark suite spanning code, instruction following, reasoning, and multi-turn agentic function-calling. Scoped to data-curation cost, the cost-adjusted trade rate $\rho_{\text{cost}}$ between gated synthetic and human-authored RLVR tasks stays in $[1.4\times, 11.5\times]$ across the plausible $c_{\text{human}}/c_{\text{aug}}$ range.
\end{abstract}

% ============================================================
\section{Introduction}
\label{sec:intro}

Hand-authoring a single agentic RLVR training task costs hours of expert time. Each task needs a Docker environment, setup scripts, a natural-language prompt, and a task-defined reward function, plus quality-assurance overhead before the task clears a learnable-zone gate. Only tasks that pass that gate (verifier correctness, learnable-zone solvability, distinctness, faithfulness) produce useful training signal, so the effective supply of high-quality training tasks is one of the central bottlenecks reinforcement learning from verifiable rewards (RLVR) on agentic language models runs against. At the scale required for effective RLVR training (hundreds to thousands of tasks), human curation alone is economically impractical. Prior work scaling RL training data has defaulted to expanding human curation, with OpenAssistant~\cite{openassistant} collecting 161k human dialogue annotations, T\"{u}lu 3~\cite{tulu3} relying on extensive human-curated post-training mixes, and SWE-Gym~\cite{swegym} hand-curating agentic GitHub-based environments. Prior synthetic-data work in this setting~\cite{resyn,agent-rlvr} has not established the cost-quality limit for augmentation, leaving expanded human curation as the de facto scaling route.

To test whether synthetic task variants can substitute for additional human curation, we run a controlled ablation that measures the trade rate $\rho$, the ratio of marginal pass@1 gain per augmented task to marginal pass@1 gain per human task. Together with a cost ratio $c_{\text{human}}/c_{\text{aug}}$, $\rho$ determines whether augmentation dominates human curation as a scaling strategy. The four primary arms follow the naming convention \emph{H$h$\_A$a$} with $h$ hand-authored tasks and $a$ augmented variants. \emph{H10\_A0} and \emph{H97\_A0} are human-only. \emph{H10\_A80} (10 base $+$ 80 augmented) is near-step-matched to H97\_A0, and \emph{H10\_A319} (10 base $+$ 319 augmented) is the cost-advantaged scaled augmented arm (cheaper than H97\_A0 across the swept $c_{\text{human}}/c_{\text{aug}}$ range). We additionally train two compute-matched extended arms, \emph{H97\_A0$^{\dagger}$} and \emph{H10\_A80$^{\dagger}$}, to the H10\_A319 total step count (330 steps) under the same configuration, isolating task source from training-compute confounds. All other GRPO hyperparameters, model (Qwen3.5-27B), evaluation schedule, and held-out suite are held fixed across arms.

\begin{figure}[ht]
\centering
\includegraphics[width=\linewidth]{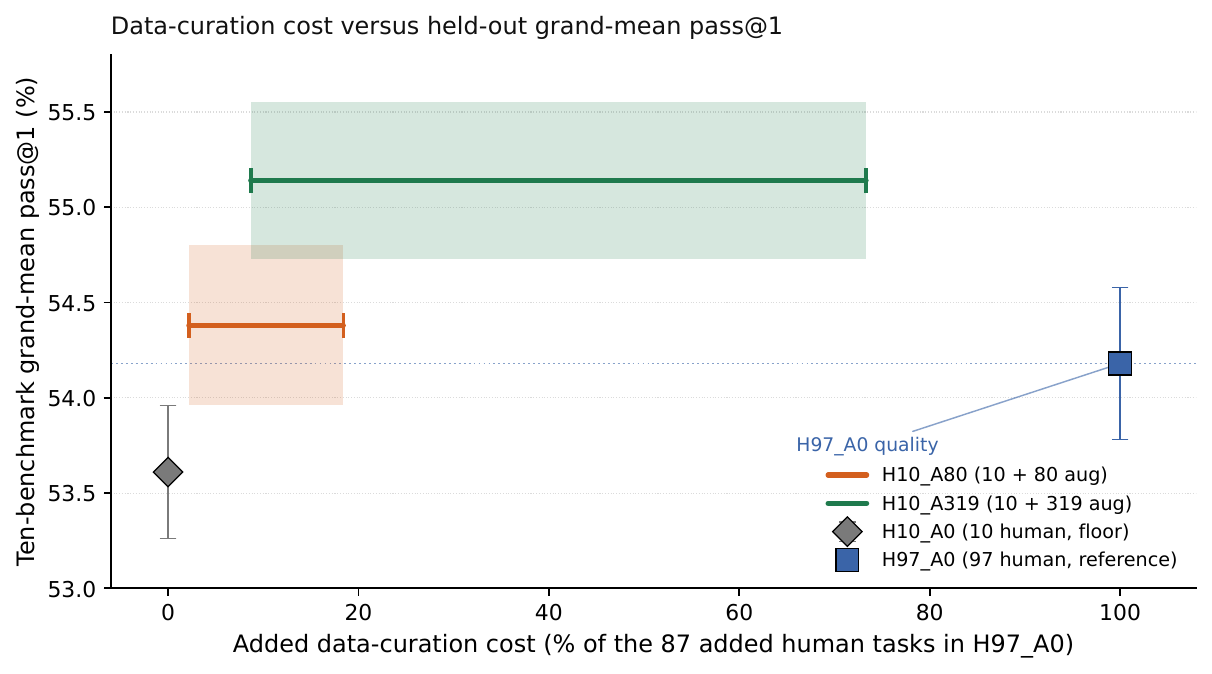}
\caption{Added data-curation cost ($x$, as a percentage of the 87 additional hand-authored tasks in H97\_A0) versus ten-benchmark grand-mean pass@1 ($y$). H10\_A0 (0\% added cost) and H97\_A0 (100\%) are exact points. The augmented arms are horizontal bands spanning the swept $c_{\text{human}}/c_{\text{aug}} \in [5\times, 42\times]$ range, plotted as $1/\rho_{\text{cost}}$ (H10\_A80 spans $2.2$--$18.4\%$, H10\_A319 spans $8.7$--$73.3\%$). Single training seed. Error bars and band heights are decoding-seed standard deviation only.}
\label{fig:headline}
\end{figure}

From a 10-task hand-authored base, 80 gated augmented variants reach the held-out generalization band of 97 fully hand-authored tasks on a ten-benchmark suite spanning code, instruction following, reasoning, knowledge, and multi-turn agentic tool use (grand-mean within $0.20$ percentage points of the human-only baseline, Figure~\ref{fig:headline}). At $4\times$ scaled augmentation, 319 augmented variants are directionally better than the 97-task human-only baseline by $+0.96$ percentage points on the aggregate, with 8 of 10 per-benchmark wins. The cost-adjusted trade rate $\rho_{\text{cost}}$ between gated synthetic and human-authored RLVR tasks stays in $[1.4\times, 11.5\times]$ across the plausible $c_{\text{human}}/c_{\text{aug}}$ range.

Our primary contribution is a pilot, single-training-seed result (three decoding seeds per held-out cell). The data are consistent with pre-specified, gate-filtered augmentation of a small hand-authored base substituting for additional human curation in agentic RLVR at lower data-curation cost, under the aggregate-parity criterion (\S\ref{sec:bench}) and the swept $c_{\text{human}}/c_{\text{aug}}$ range (\S\ref{sec:pipeline_econ}). We:
\begin{itemize}
  \item formalize the cost-adjusted trade rate $\rho_{\text{cost}}$ between synthetic and human-authored RLVR tasks and measure it end-to-end against matched compute- and near-step-matched human baselines (\S\ref{sec:experiment}, \S\ref{sec:rho});
  \item show that high-share augmentation reaches the held-out generalization band of additional human curation on a ten-benchmark held-out suite (H10\_A80 grand-mean within $0.20$ percentage points of H97\_A0; H10\_A319 directionally better at $+0.96$ percentage points, 8 of 10 per-benchmark wins), at $\rho_{\text{cost}} \in [1.4\times, 11.5\times]$ across the plausible $c_{\text{human}}/c_{\text{aug}}$ range (\S\ref{sec:bench}, Figure~\ref{fig:headline});
  \item characterize the pipeline calibration regime, with $25.5\%$ gate yield and a dominant $64\%$ \texttt{too\_easy} rejection mode that identifies planner overshoot toward easy variants as the next calibration target (\S\ref{sec:pipeline_econ}).
\end{itemize}

% ============================================================
\section{Related Work}
\label{sec:related}

\subsection{Reinforcement Learning from Verifiable Rewards (RLVR)}

Yue et al.~\cite{yue2025} argue that RLVR gains come from sampling efficiency on already-solvable problems rather than capability expansion, with pass@1 improving while pass@$k$ at large $k$ does not exceed the base model. Verifier reliability is also a known concern. TinyV~\cite{tinyv} reports false-negative rates up to 38\% on rule-based verifiers. Agentic-RLVR scaling has been demonstrated at the algorithm and model-size axes, with Agent-RLVR~\cite{agent-rlvr} lifting SWE-bench from 9.4\% to 22.4\% on a 72B model, DeepSeek-R1-Zero~\cite{deepseek-r1} showing reasoning emergence via pure RL without SFT, and Med-RLVR~\cite{med-rlvr} reporting $+8$ percentage points out-of-distribution gains on a 3B base. We respond to verifier reliability with a multiplicative quality gate (Section~\ref{sec:quality}) and hybrid verification (\S~\ref{app:verifier}). Orthogonal to these works, our contribution targets the \emph{task generation} axis of agentic RLVR, not the algorithm or model size.

\subsection{Synthetic Task Augmentation for RL}

Data augmentation is well-established in supervised learning but underexplored for RL training environments. In vision-based RL, image augmentations (crop, shift, color jitter) improve sample efficiency~\cite{rad,drq}. For language tasks, paraphrasing and backtranslation provide surface diversity without altering the underlying task. Curriculum methods~\cite{paired,plr} \emph{select} or generate tasks at the frontier of agent competence. Closest to our work, ReSyn~\cite{resyn} reports $+27\%$ on BBEH via synthetic task augmentation, AgentRL~\cite{agentrl} achieves SOTA on ALFWorld/WebShop/DB with multi-turn multi-task RL, and ProRL Agent~\cite{prorl} validates the rollout-as-a-service pattern for agentic RL. Neither ReSyn nor AgentRL formalizes a substitution rate between synthetic and human-authored tasks or reports end-to-end pipeline cost (acceptance yield, per-axis variance, per-variant dollar cost). Both demonstrate that augmentation helps without quantifying how much it substitutes per dollar. We expand the \emph{pool} of training tasks through a pre-specified mutation pipeline that changes what the task \emph{requires} the agent to do (gated by automated quality filters within the learnable zone pass@8 $\in [0.05, 0.95]$) and measure the resulting per-dollar substitution rate end-to-end against matched compute- and near-step-matched human baselines.

\subsection{Cost-Quality Trade-offs in AI Systems}

Recent work frames AI system design as a \emph{cost-quality trade-off} problem. Hybrid-LLM routing~\cite{hybrid-llm} routes $22\%$ of queries to a cheaper model with $<$$1\%$ quality drop, crowdsourcing-with-error-correction~\cite{embrace-error} achieves $10\times$ throughput at controlled error rates, and reasoning-token budgeting~\cite{token-budget} compresses without proportional performance loss. We apply the same lens to \emph{training data}, treating human tasks as the expensive, high-quality option and augmented tasks as the cheap, potentially-lower-quality option, with the trade rate $\rho$ quantifying training signal lost per dollar saved.

% ============================================================
\section{Methodology}
\label{sec:method}

\subsection{Augmentation Pipeline}
\label{sec:pipeline}

\begin{figure}[ht]
\centering
\begin{tikzpicture}[
  node distance=4mm,
  every node/.style={font=\small},
  box/.style={draw, rounded corners=1.2pt, align=center, inner sep=3pt, line width=0.4pt},
  source/.style={box, fill=gray!12},
  pipeline/.style={box, fill=blue!7},
  gate/.style={box, fill=orange!12},
  output/.style={box, fill=green!10},
  reject/.style={box, fill=red!8, font=\small, align=center, inner sep=3pt, line width=0.4pt, rounded corners=1.2pt, draw},
  arrow/.style={-{Latex[length=2mm]}, line width=0.5pt},
  rejarrow/.style={-{Latex[length=1.6mm]}, dashed, line width=0.5pt, gray}
]
  \node[source] (base) {10 base\\tasks};
  \node[pipeline, right=6mm of base] (scout) {Scout\\variant};
  \node[pipeline, right=6mm of scout] (fanout) {Mutation\\strategies};
  \node[pipeline, right=6mm of fanout] (div) {Diversity\\steering};
  \node[gate, right=6mm of div] (gate) {Gate \\$Q \!\geq\! 0.5$};
  \node[output, right=6mm of gate] (accept) {Accepted\\variant};
  \node[reject, below=8mm of gate] (rej) {Reject 74.5\%\\(64\% \texttt{too\_easy})};

  \draw[arrow] (base) -- (scout);
  \draw[arrow] (scout) -- (fanout);
  \draw[arrow] (fanout) -- (div);
  \draw[arrow] (div) -- (gate);
  \draw[arrow] (gate) -- (accept);
  \draw[rejarrow] (gate) -- (rej);
\end{tikzpicture}
\caption{Augmentation-pipeline lifecycle. Each base task expands through a scout variant, parallel pre-specified mutation strategies, and diversity steering (domain + failure-mode axes). Variants pass through the multiplicative quality gate $Q = V_{\text{verify}} \!\times\! V_{\text{solve}} \!\times\! V_{\text{distinct}} \!\times\! V_{\text{faithful}} \!\times\! V_{\text{informative}}$ at threshold 0.5.}
\label{fig:pipeline}
\end{figure}
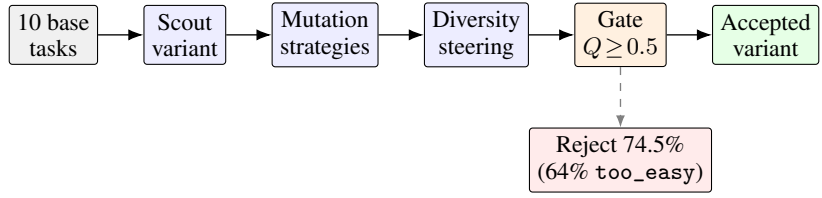

For each base task we (i) generate a \emph{scout} variant that probes task-specific constraints, then (ii) fan out remaining variants in parallel using pre-specified deterministic mutation strategies (information removal, constraint densification, reference vaguification; full enumeration in \S~\ref{app:strategies}), following the seed-task-fan-out tradition of Self-Instruct~\cite{self-instruct} and the directed-mutation tradition of Evol-Instruct~\cite{evol-instruct}. Both the augmentation engine and the agentic component of the task-defined reward function are implemented with a frontier model (Claude Opus 4.6, accessed via API; the same model fills both roles in this study). Reproducibility implications of this closed-weight dependency are discussed in \S~\ref{app:reproducibility}. Each variant is then passed through the quality gate (Section~\ref{sec:quality}). Diversity steering prevents sibling redundancy across the variant pool (\S~\ref{app:diversity}). Figure~\ref{fig:pipeline} traces the full variant lifecycle from base task through gate decision.

\subsection{Quality Gate}
\label{sec:quality}

Every variant is scored on a multiplicative gate where a single zero kills the variant:

\begin{equation}
Q = V_{\text{verify}} \times V_{\text{solve}} \times V_{\text{distinct}} \times V_{\text{faithful}} \times V_{\text{informative}}
\end{equation}

\begin{itemize}
  \item $V_{\text{verify}}$: verifier correctness, validated by known-good and known-bad solutions.
  \item $V_{\text{solve}}$: learnable-zone check, pass@8 $\in [0.05, 0.95]$, where always- or never-solved tasks produce zero GRPO gradient.
  \item $V_{\text{distinct}}$: KS test on reward distributions, rejected if indistinguishable from $>$80\% of siblings.
  \item $V_{\text{faithful}}$: solution-transfer test plus LLM-as-judge against the mutation specification.
  \item $V_{\text{informative}}$: runtime fraction of non-zero-gradient samples.
\end{itemize}

Threshold: $Q \geq 0.5$. Each task carries a task-defined reward function (\S~\ref{app:verifier}). Reward composition is held fixed across all six arms and not analyzed in this paper.

% ============================================================
\section{Experiment Design}
\label{sec:experiment}

\subsection{Economic Substitution Framing}

We frame the choice between human task authoring and automated augmentation as a substitution problem. Given two training data sources with different per-unit costs and per-unit training value, what is their substitution elasticity, and under what cost ratios does one dominate? This mirrors the cost-quality routing formulation in hybrid LLM inference~\cite{hybrid-llm}.

\subsection{Research Questions}

\begin{enumerate}
  \item \textbf{Trade rate}: how many augmented tasks equal one human task in marginal training value, both per-task on training-set canary and per-arm on held-out generalization?
  \item \textbf{Cost-adjusted substitution}: at a given data-curation cost ratio, does augmentation match or outperform human curation on held-out generalization?
  \item \textbf{Diminishing returns}: where does the marginal value of additional augmented variants drop below the cost of generating them?
  \item \textbf{Failure-mode coverage}: do augmented arms elicit a wider distribution of agent failure modes than human-only training?
\end{enumerate}

\subsection{Arm Design and Validity Controls}

\begin{table}[ht]
\centering
\caption{Ablation arms. All arms use standard GRPO with identical hyperparameters, model, LoRA config, and evaluation schedule. The only variable is the training task set. Arm names follow \emph{H$h$\_A$a$} with $h$ hand-authored tasks plus $a$ augmented variants. \emph{H10\_A0} and \emph{H97\_A0} are human-only. \emph{H10\_A80} and \emph{H10\_A319} hold the same 10-task hand-authored base and add 80 or 319 augmented variants.}
\label{tab:arms}
\begin{tabular}{@{}l>{\raggedright\arraybackslash}p{3.2cm}p{6.8cm}@{}}
\toprule
\textbf{Arm} & \textbf{Training Tasks} & \textbf{What It Isolates} \\
\midrule
H10\_A0        & 10 base human tasks & Baseline performance from minimum viable human set \\
H97\_A0        & 97 human tasks ($\sim$10$\times$ H10\_A0) & Value of scaling human tasks: the expensive option \\
H10\_A80   & 10 base + 80 augmented & At near-equal task count to H97\_A0, is the gap due to task \emph{source} or task \emph{count}? \\
H10\_A319  & 10 base + 319 augmented & Cost-advantaged scaled augmented arm: with the same base set as H10\_A80, what is the marginal-value curve under $4\times$ more variants? Cheaper than H97\_A0 across the swept $c_{\text{human}}/c_{\text{aug}}$ range. \\
\bottomrule
\end{tabular}
\end{table}

\paragraph{Isolation of effects.} Each comparison isolates exactly one variable:
\begin{itemize}
  \item \textbf{H97\_A0 vs.\ H10\_A80} isolates \emph{task source} (human vs.\ augmented) at fixed task count.
  \item \textbf{H97\_A0 vs.\ H10\_A319} isolates \emph{allocation strategy} at fixed budget.
  \item \textbf{H10\_A80 vs.\ H10\_A319} isolates \emph{scale} of augmentation at fixed base set.
  \item \textbf{H10\_A0 vs.\ H97\_A0} establishes the \emph{marginal value of one human task}.
\end{itemize}

All arms share the same GRPO hyperparameters, LoRA configuration, evaluation schedule, checkpoint cadence, and hardware. The 10 base tasks in H10\_A0, H10\_A80, and H10\_A319 are a subset of the 97 hand-authored tasks in H97\_A0, preventing confounds from task selection.

\paragraph{Quantities of interest.} The primary estimand is the quality trade rate $\rho$. For a generic augmented arm $X \in \{$\texttt{H10\_A80}, \texttt{H10\_A319}$\}$ measured against the H10\_A0 base and the H97\_A0 human reference,
\begin{equation}
\rho_X = \frac{(\text{pass@1}_X - \text{pass@1}_{\text{H10\_A0}}) / N_{\text{aug},X}}{(\text{pass@1}_{\text{H97\_A0}} - \text{pass@1}_{\text{H10\_A0}}) / N_{\text{human added}}}
\end{equation}
with $N_{\text{aug}, \texttt{H10\_A80}} = 80$, $N_{\text{aug}, \texttt{H10\_A319}} = 319$, and $N_{\text{human added}} = 87$. If $\rho_X = 0.5$, two augmented tasks produce the same marginal training value as one human task. The cost-adjusted trade rate $\rho_{\text{cost}}$ is the ratio of the human-curation cost to the augmentation-pipeline cost for two arms that match in held-out generalization:
\begin{equation}
\rho_{\text{cost}} = \frac{c_{\text{human}} \cdot N_{\text{human added}}}{c_{\text{aug}} \cdot N_{\text{aug}}}
\end{equation}
where $N_{\text{human added}}$ and $N_{\text{aug}}$ are the additional task counts beyond the shared 10-task base and $c_{\text{human}}$, $c_{\text{aug}}$ are the fully-loaded per-task costs. $\rho_{\text{cost}} > 1$ indicates the augmented arm matches or exceeds the human-only arm's held-out quality at lower curation cost. This parallels the cost-advantage threshold in hybrid inference routing~\cite{hybrid-llm}.

\paragraph{Identifying assumptions.} The estimand is the per-task substitution rate between augmented and human-authored agentic RLVR tasks, with potential outcomes defined as held-out pass@1 under each corpus assignment for an identically-initialized policy~\cite{athey-imbens-2017,saito-joachims-2022}. The unit is the trained policy at the final training checkpoint. Identification rests on three controls: (i) all arms initialize from the same base policy and train with identical hyperparameters, so cross-arm differences in held-out outcomes are attributable to training-corpus content; (ii) the 10-task base in H10\_A0, H10\_A80, and H10\_A319 is the same subset of the 97 hand-authored tasks in H97\_A0, eliminating task-selection confound at the base; (iii) compute-matched extended arms (H97\_A0$^{\dagger}$, H10\_A80$^{\dagger}$) train to the H10\_A319 total step count under the same configuration, isolating task-source effects from training-compute confounds. Identification breaks under judge-API drift across arms, held-out task leakage, or systematic eval-distribution shift. We hold the judge model version fixed, build the held-out suite from public benchmarks unused at training, and keep the eval suite identical across arms.

\subsection{Setup}
\label{sec:setup}

Tasks for H10\_A0 and H97\_A0 are drawn from a curated agentic-task catalog spanning data science, scientific computing, software engineering, gitops, and sysops domains. Each task includes a Docker environment, setup scripts, a natural-language prompt, and a task-defined reward function. H10\_A80 and H10\_A319 use the 10 base tasks from H10\_A0 plus pipeline-generated variants (80 accepted in H10\_A80, 319 accepted in H10\_A319), with no additional human-authored tasks. All evaluation benchmarks are held out from training.

All six arms train Qwen3.5-27B with GRPO at identical per-step compute and identical hyperparameters (LoRA rank 16, $\alpha=32$, lr $10^{-4}$, group size 4, max 30 turns/episode, 4{,}096 tokens, KL coef 0.001) on SageMaker p4d.24xlarge. The four primary arms (H10\_A0, H97\_A0, H10\_A80, H10\_A319) train for 4 epochs over their respective corpora, so total step count scales with corpus size: H97\_A0 and H10\_A80 land at near-equal step counts (roughly 97 and 92 steps), while H10\_A319 reaches 330 steps at 4 epochs over its larger corpus. The two compute-matched extended arms, H97\_A0$^{\dagger}$ and H10\_A80$^{\dagger}$, train their smaller corpora past 4 epochs to the same 330-step budget, isolating task source from training-compute confounds. Rollouts use an in-house harness that runs Docker-isolated episodes through a LiteLLM proxy capturing per-token logprobs to S3, and LoRA updates hot-swap to the SageMaker endpoint without restart. Full hyperparameter and infrastructure detail in \S~\ref{app:setup}, and compute budget in \S~\ref{app:compute}.

% ============================================================
\section{Evaluation}
\label{sec:eval}

\subsection{Benchmark Suite}

We evaluate all six trained checkpoints (four primary arms plus the two compute-matched extended arms H97\_A0$^{\dagger}$ and H10\_A80$^{\dagger}$) on a held-out suite spanning three transfer distances from the training distribution and two task formats. The \emph{in-domain} (single-turn) benchmarks are HumanEval and DS-1000. \emph{Near-domain} (single-turn) benchmarks are IFEval (instruction following) and MATH500 (reasoning). \emph{Far-domain} (single-turn) benchmarks are BBEH, MMLU-Pro, and GPQA Diamond. The \emph{held-out agentic} (multi-turn) benchmarks are BFCL v4 (Berkeley Function-Calling Leaderboard v4, overall accuracy), BFCL multi-turn (the BFCL v4 multi-turn category), and $\tau^2$-bench retail (Sierra dual-control retail customer-service simulator). The three agentic benchmarks are the most direct test of generalization for a multi-turn-trained policy.

\begin{table}[ht]
\centering
\caption{Evaluation benchmark suite. Each benchmark is evaluated at the final training checkpoint per arm with pass@1 reported.}
\label{tab:benchmarks}
\small
\begin{tabular}{@{}llllr@{}}
\toprule
\textbf{Benchmark} & \textbf{Capability} & \textbf{Ring} & \textbf{Format} & \textbf{$n$} \\
\midrule
HumanEval~\cite{chen2021codex} & Code generation & In & Single-turn & 164 \\
DS-1000~\cite{ds1000} & Data-science code gen & In & Single-turn & 1{,}000 \\
\midrule
IFEval~\cite{ifeval} & Instruction following & Near & Single-turn & 541 \\
MATH500~\cite{math500} & Math reasoning & Near & Single-turn & 500 \\
\midrule
BBEH~\cite{bbeh} & Reasoning (23 subtasks) & Far & Single-turn & 200 \\
MMLU-Pro~\cite{mmlu_pro} & Knowledge \& reasoning & Far & Single-turn & 1{,}000 \\
GPQA Diamond~\cite{gpqa} & Graduate science Q\&A & Far & Single-turn & 198 \\
\midrule
BFCL v4~\cite{bfcl} & Function-calling tool use & Agentic (held-out) & Multi-turn & $\sim$200 \\
BFCL multi-turn~\cite{bfcl} & Multi-turn function calling & Agentic (held-out) & Multi-turn & $\sim$200 \\
$\tau^2$-bench retail~\cite{tau2bench} & Retail customer-service tool use & Agentic (held-out) & Multi-turn & 114 \\
\bottomrule
\end{tabular}
\end{table}

The suite spans code (HumanEval, DS-1000), instruction following (IFEval), math reasoning (MATH500), knowledge and reasoning (BBEH, MMLU-Pro, GPQA Diamond), and three multi-turn agentic benchmarks (BFCL v4, BFCL multi-turn, $\tau^2$-bench retail) closest in format to the training distribution. All ten benchmarks are held out from training by task content.

\subsection{Metrics and Statistical Reporting}

The headline substitution claim is supported by held-out pass@1 (\S\ref{sec:bench}) plus the $c_{\text{human}}/c_{\text{aug}}$ cost-sensitivity sweep (\S\ref{sec:pipeline_econ}). Training-set fit on the canary (\S\ref{sec:rho}) is a diagnostic, not the primary substitution evidence. We report \textbf{pass@1} on each held-out benchmark across three decoding seeds per cell ($n=2$ for BFCL multi-turn) as mean $\pm$ std. The primary metric is the ten-benchmark grand-mean aggregate per arm. Per-benchmark $z$-tests are reported for context, but given the within-benchmark sampling-noise width of $\pm 1$ to $\pm 5$ percentage points, no single-arm-pair test in the H10\_A319 vs.\ H97\_A0 family clears $\alpha=0.05$ uncorrected and no Holm--Bonferroni adjustment changes the conclusion (\S\ref{sec:bench}). Per-run JSONs for all clean runs are released in supplementary material.

\paragraph{Harnesses.} HumanEval, DS-1000 $\to$ bigcode-eval-harness with sandboxed code execution; IFEval, MMLU-Pro, MATH500, GPQA $\to$ lm-eval-harness; BBEH $\to$ Inspect AI; BFCL v4, BFCL multi-turn, and $\tau^2$-bench retail (all three held-out agentic) $\to$ the same in-house rollout harness used during training, with serving via the trained-checkpoint endpoints, and for $\tau^2$-bench retail the public Sierra simulator (\texttt{sierra-research/tau2-bench}) for the user model and dual-control environment. Serving and harness configuration in \S~\ref{app:eval}.

\paragraph{Training--evaluation distinctness.} No held-out benchmark prompts, gold answers, or test-set inputs appear in any training arm by construction. Training is multi-turn agentic Docker-isolated tasks while held-out evaluation spans static single-turn benchmarks plus the BFCL v4 and BFCL multi-turn function-calling evaluations and Sierra's $\tau^2$-bench dual-control simulator. Pretrained-checkpoint contamination of HumanEval is plausible at the 92.7\% base, so we treat HumanEval as a saturation-zone secondary check and rely on contamination-resistant DS-1000, GPQA Diamond, and BFCL v4 as the load-bearing in-domain and far-domain benchmarks.

% ============================================================
\section{Results}
\label{sec:results}

\subsection{Training Dynamics and Canary Diagnostics}
\label{sec:rho}

As a training-set fit diagnostic we estimate $\rho_X$ (defined in Section~\ref{sec:experiment}) on the 10 base tasks present in all four primary arms (the canary set), evaluated at the final training step (Table~\ref{tab:exchange_rate}). The canary set is in every arm's training data, so the resulting $\rho_C \equiv \rho_{\texttt{H10\_A80}}$ and $\rho_D \equiv \rho_{\texttt{H10\_A319}}$ measure training-set fit, not generalization. The headline substitution claim depends on held-out generalization (\S\ref{sec:bench}, Table~\ref{tab:bench_main}), not on this number.

\begin{table}[ht]
\centering
\caption{Final-step pass@1 on the 10 shared base tasks. $\rho$ measures marginal pass@1 gain per augmented task relative to per human task. Cost-adjusted comparison and $\rho_{\text{cost}}$ derivation in \S\ref{sec:pipeline_econ}.}
\label{tab:exchange_rate}
\small
\begin{tabular}{@{}lcccc@{}}
\toprule
\textbf{Quantity} & \textbf{H10\_A0} & \textbf{H97\_A0} & \textbf{H10\_A80} & \textbf{H10\_A319} \\
& \textbf{(10 base)} & \textbf{(97 human)} & \textbf{(10+80 aug.)} & \textbf{(10+319 aug.)} \\
\midrule
final-step pass@1 (shared base) & 0.063 & 0.125 & 0.100 & 0.113 \\
$N_{\text{tasks added}}$ vs.\ A & 0 & +87 (human) & +80 (aug.) & +319 (aug.) \\
data-curation cost (rel.\ to H10\_A319) & --- & $1.4\times$--$11.5\times$ & $0.25\times$ & $1\times$ \\
\midrule
$\rho$ on pass@1 & --- & 1.00 (def.) & 0.65 & 0.22 \\
\bottomrule
\end{tabular}
\end{table}

\paragraph{Reading the result.} On the shared canary, the training-set fit ratio is $\rho_C = 0.65$ at near-compute parity (H10\_A80 vs.\ H97\_A0) and $\rho_D = 0.22$ at scaled augmentation (H10\_A319 vs.\ H97\_A0). $\rho_D / \rho_C \approx 1/3$ marks a diminishing-returns elbow on the canary where scaling augmentation $4\times$ at fixed base retrieves only $1.3\times$ the pass@1 gain. We surface these as training-set diagnostics. They are not the substitution rate that the headline cost-dominance claim (\S\ref{sec:bench}, Table~\ref{tab:bench_main}) depends on.

\begin{figure}[ht]
\centering
\includegraphics[width=\linewidth]{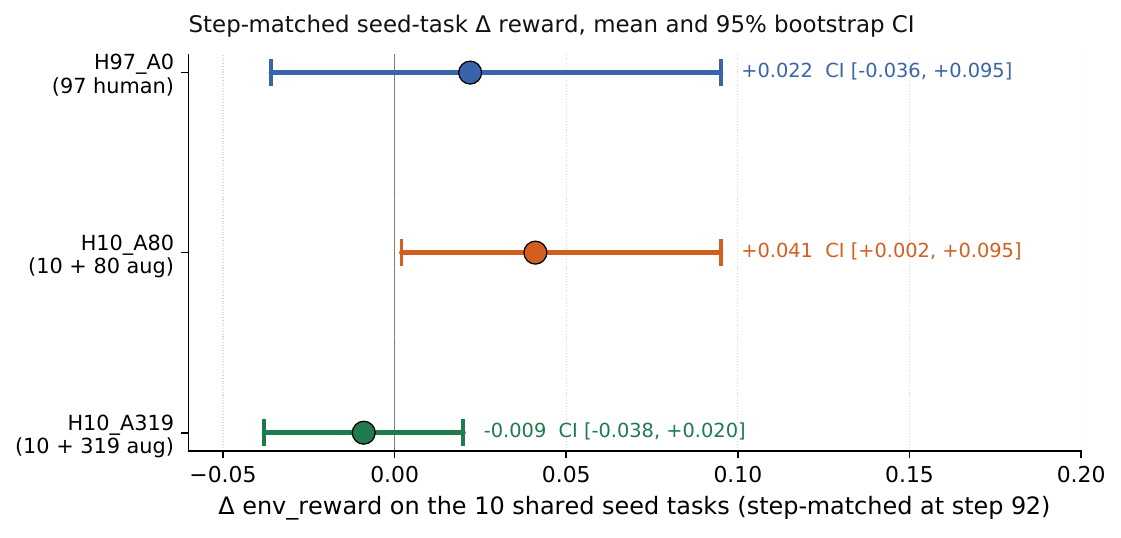}
\caption{Step-matched ($\leq$$92$ steps) seed-task $\Delta$ env\_reward per arm, mean and $95\%$ bootstrap CI across the 10 shared base tasks. H10\_A80 is the only arm whose CI is strictly above zero. H97\_A0 and H10\_A319 both cross zero. H10\_A80 wins on the canary at the same compute budget as the human-only control.}
\label{fig:matched_compute}
\end{figure}

\paragraph{Step-matched seed-task lift.} To rule out the ``H10\_A80 simply ran longer'' attack, we cut all three arms at the H10\_A80 step budget (step $92$) and recompute the per-seed $\Delta$ env\_reward lift (raw, $[0,1]$ scale) on the 10 shared base tasks with $95\%$ bootstrap CIs. H10\_A80's mean lift is $+0.041$ with CI $[+0.002, +0.095]$ (strictly above zero); H97\_A0's $+0.022$ CI $[-0.036, +0.095]$ crosses zero; H10\_A319's $-0.009$ CI $[-0.038, +0.020]$ crosses zero (D needs its full budget to reach its outcome). H10\_A80 wins at compute parity (Figure~\ref{fig:matched_compute}).

\subsection{Pipeline Economics and the Calibration Regime}
\label{sec:pipeline_econ}

The $\rho_{\text{cost}}$ headline depends on $c_{\text{aug}}$, which we measure end-to-end across the full pipeline funnel. Across the two task-sets used in H10\_A80 and H10\_A319, the planner produced 930 attempts; 237 cleared the gate (yield 25.5\%); 64\% of rejections were classified \texttt{too\_easy} (planner overshoots toward easier variants); 52\% of attempts are retries. Per-axis yield varies $6\times$, from 7.7\% (\texttt{OF12}) to 45.5\% (\texttt{DK17.1}). Rebalancing the per-axis budget toward higher-yield axes is the next calibration target (operating rules in \S\ref{app:decisions}). Across the plausible per-task hand-curation cost range, $c_{\text{human}}/c_{\text{aug}}$ runs from roughly $5\times$ at the OpenAssistant-scale dialogue-collection lower end~\cite{openassistant} to over $40\times$ at the SWE-Gym-scale hand-curated agentic-environment upper end~\cite{swegym}. Combining this with $N_{\text{human added}}/N_{\text{aug}} = 87/319$ at matched held-out quality gives $\rho_{\text{cost}} \in [1.4\times, 11.5\times]$ across the range (Table~\ref{tab:cost_analysis}). At the OpenAssistant low end of $c_{\text{human}}/c_{\text{aug}} \approx 5\times$, the 80-augmented vs.\ 87-human comparison at matched held-out quality gives a $(87/80) \cdot 5 \approx 5.4\times$ data-curation cost ratio. Midrange ratios give larger values. $c_{\text{aug}}$ excludes judge-API and augmentation-engine inference cost (closed-API frontier model, $\sim$\$0.05 per accepted variant). Folding this in at the OpenAssistant-low end raises $c_{\text{aug}}$ by an estimated 8\%, shifting low-end $\rho_{\text{cost}}$ from $1.4\times$ to $\sim 1.3\times$ but leaving cost-dominance direction unchanged across the swept range. The break-even cost ratio at which $\rho_{\text{cost}} = 1$ is $c_h/c_a \approx 3.67\times$, just below the literature low-end estimate of $5\times$.

\begin{table}[ht]
\centering
\caption{Cost-adjusted trade rate $\rho_{\text{cost}}$ for the cost-advantaged scaled arm (H10\_A319 vs.\ H97\_A0), swept across the plausible $c_{\text{human}}/c_{\text{aug}}$ range. With $N_{\text{human added}}=87$ and $N_{\text{aug}}=319$ at matched held-out quality, $\rho_{\text{cost}} = (87/319) \cdot (c_{\text{human}}/c_{\text{aug}})$.}
\label{tab:cost_analysis}
\small
\begin{tabular}{@{}lccccc@{}}
\toprule
$c_{\text{human}}/c_{\text{aug}}$ & $5\times$ & $10\times$ & $20\times$ & $30\times$ & $42\times$ \\
$\rho_{\text{cost}}$ & $1.4\times$ & $2.7\times$ & $5.5\times$ & $8.2\times$ & $11.5\times$ \\
\bottomrule
\end{tabular}
\end{table}

\subsection{Held-out Generalization}
\label{sec:bench}

\begin{figure*}[ht]
\centering
\includegraphics[width=\textwidth]{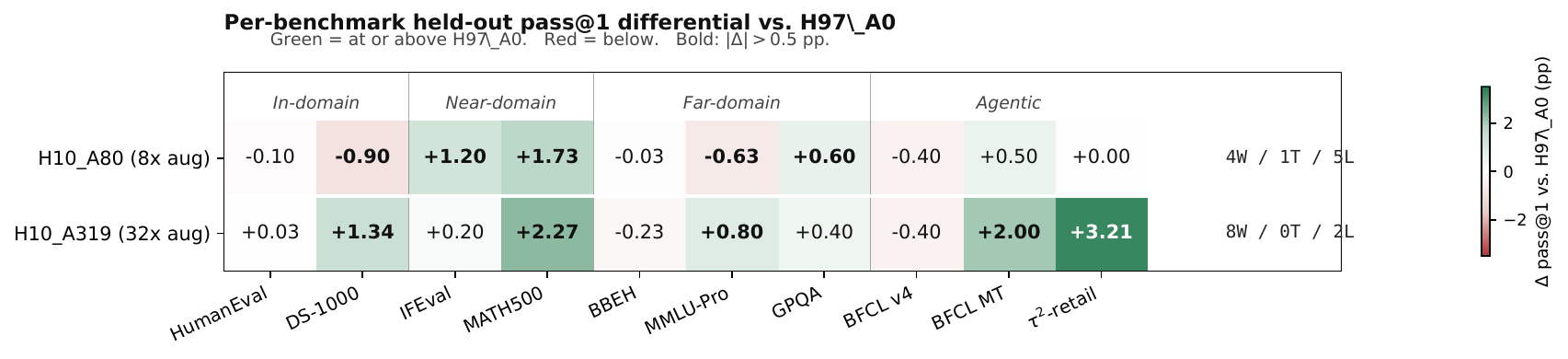}
\caption{Per-benchmark held-out pass@1 differential versus the 97-task hand-authored baseline H97\_A0, in percentage points. Rows: H10\_A80 ($8\times$ augmented) and H10\_A319 ($32\times$ augmented). Columns: ten held-out benchmarks grouped by ring (in-domain, near-domain, far-domain, agentic). Green cells mark the augmented arm at or above H97\_A0, red below; bold marks $|\Delta| > 0.5$ percentage points. Per-arm win/tie/loss tallies are shown at right.}
\label{fig:bench_delta_heatmap}
\end{figure*}

We evaluate the four primary arms and two compute-matched extended arms (H97\_A0$^{\dagger}$, H10\_A80$^{\dagger}$ trained to the H10\_A319 step count) on a ten-benchmark held-out suite spanning code (HumanEval, DS-1000), instruction following (IFEval), reasoning (MATH500), knowledge (BBEH, MMLU-Pro, GPQA Diamond), and multi-turn agentic tool use (BFCL v4, BFCL multi-turn, $\tau^2$-bench retail). Held-out evaluation is repeated across three decoding seeds per cell ($n=2$ for BFCL multi-turn). Reported numbers are mean $\pm$ std. The primary substitution metric is the ten-benchmark grand-mean per arm, reported in Table~\ref{tab:bench_main} and visualized per benchmark in Figure~\ref{fig:bench_arms}. The full per-benchmark grid appears in Appendix Table~\ref{tab:bench_full}. Per-benchmark $\Delta$ pass@1 vs.\ the H97\_A0 baseline is shown as a heatmap in Figure~\ref{fig:bench_delta_heatmap}: H10\_A319 lands at or above H97\_A0 on 8 of 10 benchmarks, H10\_A80 on 5 of 10 (with one tie).

\begin{table}[ht]
\centering
\caption{Ten-benchmark grand-mean pass@1 (\%) per arm on a single training seed; the $\pm$ values are decoding-seed variance only. H10\_A319 is the top arm and H10\_A80 reaches the H97\_A0 band within $0.20$ percentage points. Per-benchmark detail in Figure~\ref{fig:bench_arms}. Full grid in Appendix Table~\ref{tab:bench_full}.}
\label{tab:bench_main}
\small
\begin{tabular}{@{}lc@{}}
\toprule
\textbf{Arm} & \textbf{Grand-mean pass@1 (\%)} \\
\midrule
H10\_A0 \emph{(10 base, floor)}                   & $53.61 \pm 0.35$ \\
H97\_A0 \emph{(97 human, reference)}              & $54.18 \pm 0.40$ \\
H10\_A80 \emph{(10+80 aug)}                       & $54.38 \pm 0.42$ \\
H10\_A319 \emph{(10+319 aug)}                     & $\mathbf{55.14 \pm 0.41}$ \\
\midrule
H97\_A0$^{\dagger}$ \emph{(compute-matched)}      & $54.72 \pm 0.42$ \\
H10\_A80$^{\dagger}$ \emph{(compute-matched)}     & $54.11 \pm 0.29$ \\
\bottomrule
\end{tabular}
\end{table}

\begin{figure}[ht]
\centering
\includegraphics[width=\linewidth]{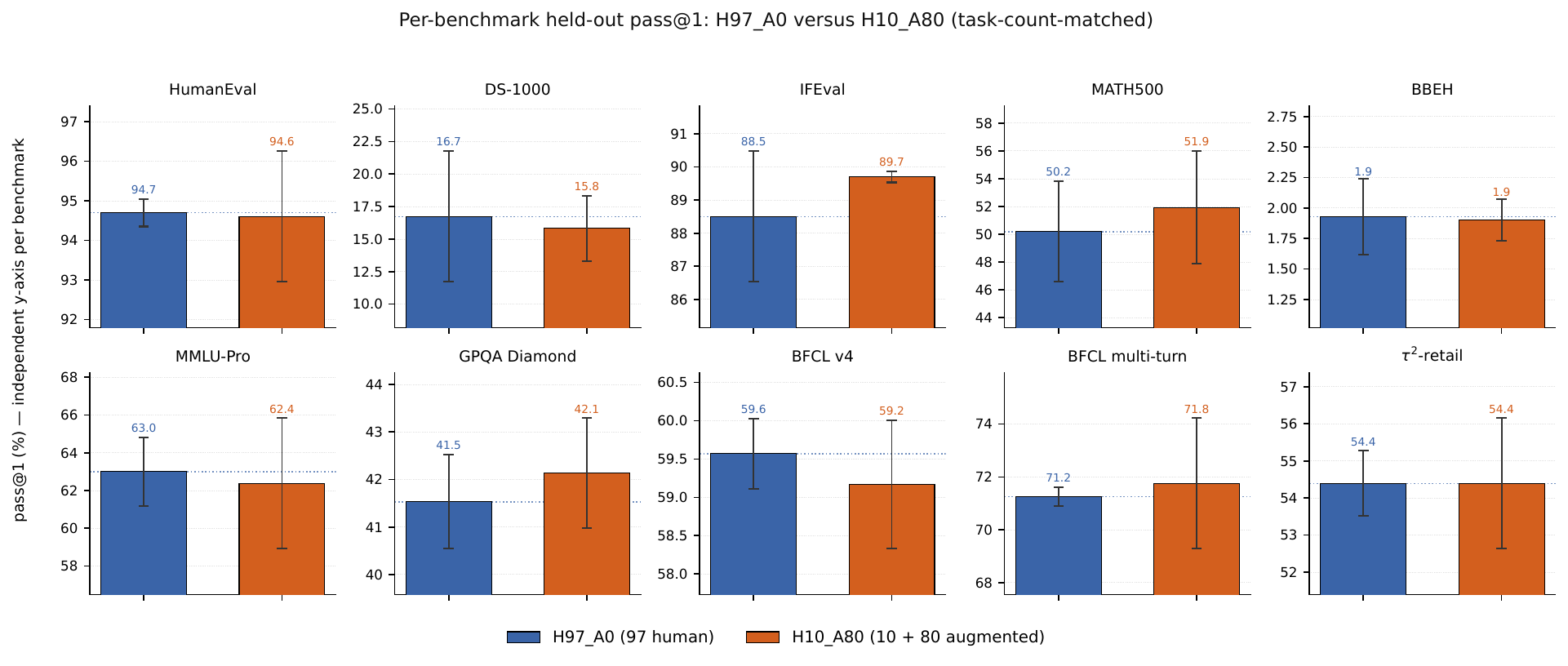}
\caption{Primary near-step-matched comparison (H10\_A80 vs.\ H97\_A0), faceted by held-out benchmark. Each panel uses an independent y-axis zoomed to that benchmark's pass@1 range so per-benchmark trends are visible without inter-benchmark scale distortion. Error bars: $\pm$ standard deviation across three decoding seeds per cell ($n=2$ for BFCL multi-turn). Dotted line: H97\_A0 reference.}
\label{fig:bench_arms}
\end{figure}

\begin{figure}[ht]
\centering
\includegraphics[width=\linewidth]{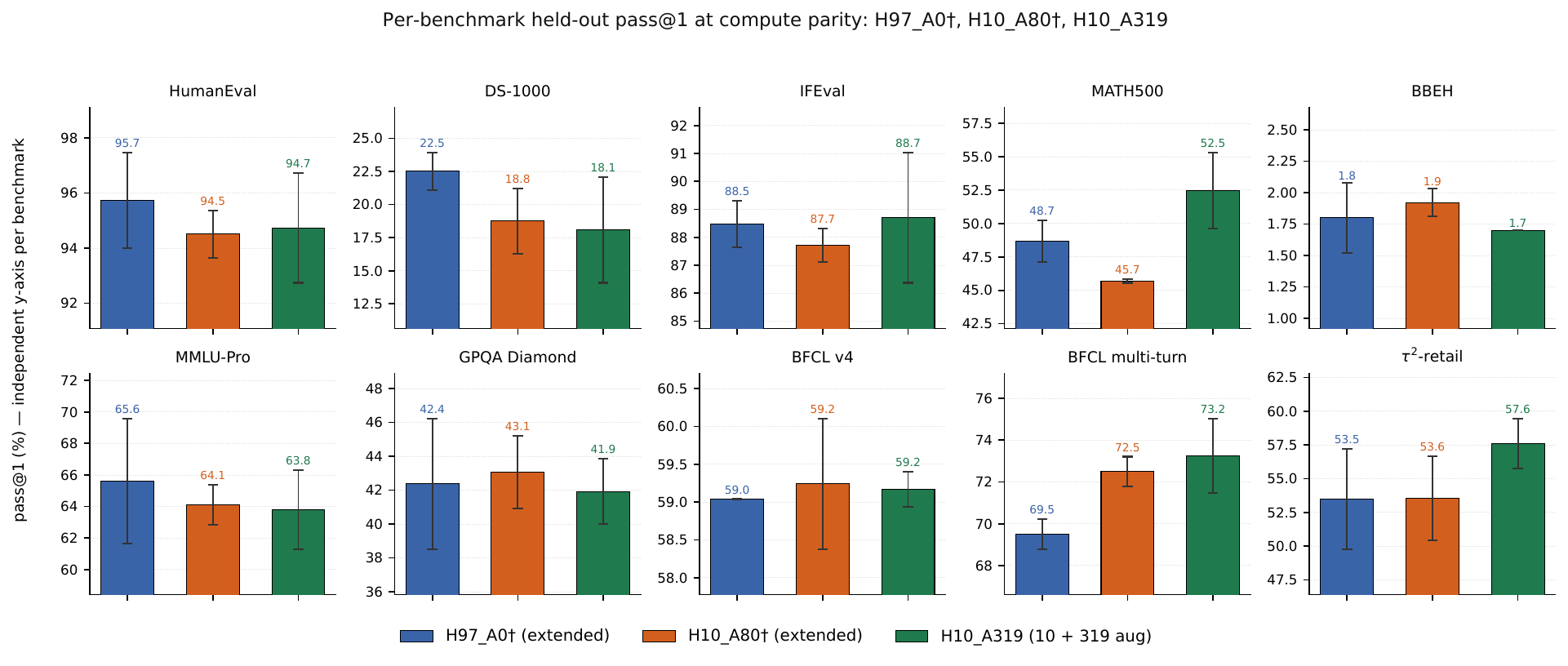}
\caption{Compute-matched comparison across all ten held-out benchmarks, faceted by benchmark. The extended human arm H97\_A0$^{\dagger}$ and extended augmented arm H10\_A80$^{\dagger}$ are trained to the H10\_A319 total step count under the same configuration, isolating task source from training-compute confounds. Each panel uses an independent $y$-axis zoomed to that benchmark's pass@1 range. Error bars: $\pm$ standard deviation across three decoding seeds per cell ($n=2$ for BFCL multi-turn).}
\label{fig:dagger}
\end{figure}

\paragraph{Aggregate held-out performance.} The ten-benchmark grand-mean (Table~\ref{tab:bench_main} bottom row) is the primary substitution metric. We equal-weight across benchmarks to avoid letting large single-turn suites (DS-1000 $n{=}1{,}000$, MMLU-Pro $n{=}1{,}000$) dominate the aggregate, treating each benchmark as one observation of held-out generalization. Augmented data reaches the performance band of additional human curation: H10\_A80 lands at $54.38$, within $+0.20$ percentage points of the 97-task hand-authored baseline H97\_A0 ($54.18$). This is the aggregate-parity result: 80 augmented variants generated from a 10-task hand-authored seed set are consistent with practical equivalence to 87 additional hand-authored tasks, though we do not establish formal non-inferiority (no pre-registered margin). H10\_A319 reaches $55.14$, $+0.96$ percentage points over H97\_A0. While directionally better, this gap is within propagated aggregate standard error (\S\ref{sec:stat_signal}). The compute-matched extended arms (H97\_A0$^{\dagger}$ $54.72$, H10\_A80$^{\dagger}$ $54.11$) preserve the same ordering at matched total step count.

\paragraph{Per-benchmark distribution.} The aggregate parity is distributional, not single-benchmark-driven. H10\_A80 stays within $\pm 1$ percentage point of H97\_A0 on 8 of 10 benchmarks and within $\pm 2$ percentage points on all 10. H10\_A319 sits at or above H97\_A0 on 8 of 10 benchmarks (the two exceptions are BBEH $-0.23$ and BFCL v4 $-0.40$, both within decoding-seed std). Augmented arms do not concentrate their performance on a narrow slice of the suite.

\paragraph{Statistical signal at aggregate scale.}
\label{sec:stat_signal}
Aggregate standard errors (propagated from per-benchmark decoding-seed std) are $\pm 0.29$ to $\pm 0.42$ per arm. These reflect decoding-seed variance only, since training is single-seed. The H10\_A80 vs.\ H97\_A0 grand-mean gap is $+0.20$ percentage points, well within aggregate noise. Aggregate parity is the supported claim. The H10\_A319 vs.\ H97\_A0 grand-mean gap is $+0.96$ percentage points (one-sided $p \approx 0.047$ via propagated aggregate SE; sign test across 10 benchmarks $p \approx 0.055$ for 8/10 wins), borderline-significant and best read as directional rather than as established superiority. Per-benchmark $z$-tests do not individually clear $\alpha=0.05$ uncorrected. Under the within-benchmark sampling noise of $\pm 1$ to $\pm 5$ percentage points, single-benchmark differences sit inside noise even at the strongest cells. The defensible claim shape is aggregate parity at near-equal training steps plus directional H10\_A319 upside, with the extra-epoch asymmetry across task source (\S\ref{sec:discussion}) as a directional follow-up question. We do not establish formal non-inferiority. Multi-seed training replication and a pre-registered non-inferiority margin are the highest-priority extensions (\S\ref{sec:limitations}). The aggregate is robust to single-benchmark removal. Under leave-one-benchmark-out, the H10\_A80 gap to H97\_A0 stays in $[+0.03, +0.32]$ percentage points across all ten drops and H10\_A319 stays above H97\_A0 in all ten (worst case $+0.71$ when the largest single contributor, $\tau^2$-bench retail, is removed). The marginal value of the 87 additional hand-authored tasks is itself small on this suite, with H97\_A0 exceeding the 10-task floor H10\_A0 by only $+0.57$ percentage points on the grand mean, so the substitution claim is scoped to this low-marginal-value regime. Augmentation matches and modestly exceeds the observed human-curation increment rather than a high-value curation ceiling.

% ============================================================
\section{Discussion}
\label{sec:discussion}

We do not claim synthetic augmentation universally dominates human curation. The defensible claim is that gated augmentation produces training signal at human-expanded performance, with directional upside at scale.

\paragraph{Aggregate ordering and compute-matched comparison.} Across the held-out suite, the aggregate ordering is H10\_A319 $>$ H97\_A0$^{\dagger}$ $>$ H10\_A80 $>$ H97\_A0 $>$ H10\_A80$^{\dagger}$ $>$ H10\_A0. Augmentation matches additional human curation on aggregate (H10\_A80 vs.\ H97\_A0 within $0.20$ percentage points) and lands modestly above it under scaled augmentation (H10\_A319 vs.\ H97\_A0, $+0.96$ percentage points, within propagated aggregate standard error). The compute-matched extended pair H97\_A0$^{\dagger}$ and H10\_A80$^{\dagger}$ trains to the H10\_A319 total step count under the same configuration, isolating task source from training compute (Table~\ref{tab:bench_main}). H10\_A319 remains the top aggregate arm at compute parity ($+0.42$ percentage points over H97\_A0$^{\dagger}$ and $+1.04$ over H10\_A80$^{\dagger}$ on the ten-benchmark grand mean). Both deltas are directional but within propagated aggregate standard error. The substitution claim therefore survives a step-count-controlled comparison.

\paragraph{Extra-epoch asymmetry across task source.} The dagger pair shows a directional asymmetry. Extending H97\_A0 to compute parity (\textasciitilde13 epochs over the same 97-task corpus) lifts grand-mean pass@1 by $+0.54$ percentage points to $54.72$, while extending H10\_A80 to compute parity (\textasciitilde14 epochs over the same 90-task corpus) moves it by $-0.27$ percentage points to $54.11$. Both shifts are within propagated aggregate standard error and individually inconclusive, but the sign asymmetry between the two arms persists. Re-cycling the human-task corpus continues to help, while re-cycling the augmented corpus does not. One reading is lower per-task information density in the augmented pool, with the gate-rejection-mode distribution ($64\%$ \texttt{too\_easy}; \S\ref{sec:pipeline_econ}) as the proximate driver. This is a directional observation only and merits multi-seed replication. We flag it as the most actionable follow-up question for practitioners deciding which corpus to scale under fixed compute.

\paragraph{Cost-dominance and the diminishing-returns elbow.} The near-step-matched augmented arm (H10\_A80) matches H97\_A0 in the aggregate-parity sense (\S\ref{sec:bench}) at lower data-curation cost across the swept $c_{\text{human}}/c_{\text{aug}}$ range (Table~\ref{tab:cost_analysis}). The canary fit ratio $\rho_D / \rho_C \approx 1/3$ marks a diminishing-returns elbow tied to the dominant \texttt{too\_easy} gate-rejection mode (\S\ref{sec:pipeline_econ}).

\paragraph{Failure-mode coverage.} On the failure-mode-coverage question of \S\ref{sec:experiment}, equal-rollout normalization ($n=472$) leaves the augmented arms exposing only $0$--$2$ additional model-side failure modes beyond the human-only arm, within bootstrap variance (\S\ref{app:failure_modes}). Augmentation does not measurably broaden the agent failure-mode distribution at this scale, so we make no capability-coverage claim; the substitution result rests on matched held-out generalization rather than on wider behavioural coverage.

\subsection{Limitations}
\label{sec:limitations}
\paragraph{Statistical limits.} Each arm is trained with a single seed. We do not claim equivalence, only no observed degradation at single-seed resolution within evaluation noise, with multi-seed training replication the highest-priority extension. Held-out evaluation is multi-seed (three decoding seeds per cell; $n=2$ for BFCL multi-turn) but training is not, so the reported CIs reflect decoding-seed variance only and do not capture training-seed variance. A negative-control intervention (re-train one arm on shuffled or sign-flipped rewards holding the rollout corpus fixed) is the next planned validation. If pass@1 degrades materially under randomized rewards, the setup is signal-sensitive to the substitution-direction reading. If it does not, the experiment is inconclusive at the current step budget.

\paragraph{Scope and cost limits.} Pilot scale (10--319 tasks per arm), a single model (Qwen3.5-27B), a single RL algorithm (GRPO), and a data-science-heavy base set leave open generalization to 10K+ corpora, 70B+ models, PPO/REINFORCE/DPO, and systematic far-domain coverage. Seven of ten held-out benchmarks are single-turn. The three multi-turn agentic benchmarks are closer matches to the training distribution, with $\tau^2$-bench airline, $\tau^2$-bench telecom, and Docker-based held-outs at SWE-bench-Verified scale deferred under harness-completeness constraints. $c_{\text{human}}/c_{\text{aug}}$ is a literature-anchored sensitivity sweep. Ground-truth $c_{\text{human}}$ on the same catalog is left to follow-up. The claim is scoped to data-curation cost only -- training compute, evaluation compute, and judge-API cost are not folded into $\rho_{\text{cost}}$. The fixed 10-task base subset and the multiplicative quality gate are validated by per-component design (\S\ref{sec:quality}) but not by resampled-base sensitivity or direct false-positive/false-negative audit, both near-term extensions. HumanEval is saturated and plausibly contaminated, so DS-1000, GPQA Diamond, and BFCL v4 are the contamination-resistant load-bearing benchmarks. End-to-end reproduction is further constrained by a proprietary task catalog, an unreleased augmentation pipeline and quality-gate scorer, unreleased LoRA adapters, and a closed frontier-model dependency for both the augmentation engine and the reward judge. We release per-arm logs, held-out benchmark outputs, and three representative augmented tasks to support partial replication.

% ============================================================
\section{Conclusion}
\label{sec:conclusion}

Starting from only 10 human seed tasks, $8\times$ gated augmentation reaches the held-out generalization band of additional human curation (grand-mean within $0.20$ percentage points of the 97-task baseline). At $32\times$ augmentation, the scaled arm is directionally better ($+0.96$ percentage points aggregate, $8$ of $10$ per-benchmark wins), though within propagated aggregate standard error. Compute-matched extended controls preserve the ordering. The cost-adjusted trade rate stays in $\rho_{\text{cost}} \in [1.4\times, 11.5\times]$. The pipeline operates in a calibration regime ($25.5\%$ acceptance yield, $64\%$ \texttt{too\_easy}-dominant rejection). The $4\times$ scale-up from H10\_A80 to H10\_A319 yields diminishing returns on training-set fit (canary $\rho_D / \rho_C \approx 1/3$). Within the swept cost range, gate-filtered augmentation substitutes for additional hand-curation at lower data-curation cost, with the strength of the claim bounded by aggregate parity and a single training seed. Lower marginal cost of agentic training-task supply enables alignment work at smaller budgets but accelerates capability proliferation under uncertain verifier reliability~\cite{tinyv}. We mitigate via training-time gating, hybrid verification (\S~\ref{app:verifier}), and no model release.

% ============================================================

% ============================================================
\appendix

\noindent\textit{Technical appendices.}
The sections below collect full hyperparameters, training and evaluation infrastructure, augmentation and verification detail, quality-gate operating decisions, failure-mode methodology, per-arm training curves, and supplementary figures referenced from the main text. Per the NeurIPS 2026 submission format, these do not count toward the nine-page main-text limit.

\section{Reproducibility and LLM Usage}
\label{app:reproducibility}

\paragraph{Release inventory and non-release rationale.}
We release per-arm training-time JSONL logs, per-arm held-out benchmark output JSONs, aggregate task-catalog statistics, analysis artifacts, Python scripts for benchmark and training-time analysis, and three representative augmented tasks (Docker environment, prompt, and task-defined reward wiring as produced by the pre-specified augmentation pipeline) as supplementary material (CC~BY-NC 4.0), with a \texttt{README.md} documenting artifact schemas, dependencies, how to reproduce tables and figures from the released logs, and how the three tasks illustrate the training task schema; hyperparameters and evaluation protocols are in \S\ref{app:setup}--\S\ref{app:eval}. Trained model checkpoints (LoRA adapters on Qwen3.5-27B) are not released. The augmentation pipeline source and quality-gate scorer are proprietary to the authors' institution and are not released; their methodology is fully specified in \S\ref{app:strategies} (augmentation strategies), \S\ref{app:diversity} (diversity steering), \S\ref{app:verifier} (verification framework), and \S\ref{app:reward} (dense reward structure) at sufficient detail to support reimplementation against any user-supplied task catalog of equivalent schema. All evaluation benchmarks are public at canonical splits and Qwen3.5-27B is public under Apache~2.0. The full base and human-expanded task catalogs remain proprietary; aside from the three illustrative augmented tasks in this supplementary bundle and aggregate catalog statistics, no additional proprietary task content is released.

\paragraph{LLM roles, API dependency, and editorial-assistance declaration.}
A single frontier model (Claude Opus 4.6, accessed via API) fills two pipeline roles, the augmentation engine and the agentic component of the task-defined reward function (\S~\ref{app:verifier}); the trained policy (Qwen3.5-27B) is the only model whose weights are updated. The augmentation engine and reward judge therefore depend on a closed-weight API-served model, which limits reproducibility for groups without API access; an open-weight substitution at the augmentation engine and judge is left to follow-up. No LLM is used in the writing beyond standard editorial assistance.

\section{Augmentation Strategies}
\label{app:strategies}

All augmentation strategies are pre-specified and organized by direction.

\paragraph{Harder.}
\emph{Densify formatting} (remove hints, increase implicit structure),
\emph{remove format examples} (force inference of output structure),
\emph{redact column names/counts} (require discovery through exploration),
\emph{vaguify references} (ambiguous pronouns, implicit relationships),
\emph{genericize constraints} (concrete examples $\to$ abstract rules),
\emph{obscure goal} (reorder prompt, defer objective).

\paragraph{Easier.} Inverse transforms: add examples, clarify structure, provide explicit names, use concrete references.

The most reliable techniques in practice are information removal ($-70$ to $-100$ percentage points on solve rate) and structural coordination ($-60$ percentage points).

\section{Divide-and-Conquer Diversity}
\label{app:diversity}

Generating hundreds of augmentations from 10 base tasks risks redundancy. We steer generation along two perspectives: (i) \emph{domain-specific diversity} (steered) distributing variants across skill domains in the source-task catalog; (ii) \emph{failure-mode diversity} (monitored) tracked over rollout root-cause labels to ensure variants exercise distinct agent failure patterns. At each fan-out step, sub-processes receive metadata steering them toward underrepresented regions while bounding overlap with siblings.

\section{Verification Framework}
\label{app:verifier}

We require verifiers to satisfy three properties:

\paragraph{Fairness.} Accept any valid solution, including alternatives to the author's intended one. Scoring weights: outcome-based $\geq 50\%$, process-based $\leq 40\%$, formatting-based $\leq 10\%$.

\paragraph{Partial grading.} Continuous scores on $[0, 1]$ rather than binary pass/fail, enabling dense reward signals.

\paragraph{Hybrid verification.} Deterministic code checks combined with an agentic judge (frontier LLM evaluating the environment state against a rubric). The agent orchestrates deterministic scripts as ``skills'' while applying judgment to ambiguous cases.

\section{Dense Reward Structure}
\label{app:reward}

A conditional dense reward ensures gradient signal even on hard tasks:

\begin{equation}
R = \begin{cases}
0.70 \times V_{\text{verify}} & \text{if } V_{\text{verify}} < 0.1 \\[4pt]
0.05 \times V_{\text{tool}} + 0.15 \times V_{\text{code}} + 0.10 \times V_{\text{verifier}} + 0.70 \times V_{\text{verify}} & \text{if } V_{\text{verify}} \geq 0.1
\end{cases}
\end{equation}

On complete failure ($V_{\text{verify}} < 0.1$), only verification contributes. On partial success, auxiliary components provide shaping gradients while the 70\% verification weight prevents reward hacking.

\section{Training and Infrastructure Configuration}
\label{app:setup}

\begin{table}[ht]
\centering
\caption{Training hyperparameters, shared across all arms.}
\label{tab:training}
\begin{tabular}{@{}ll@{}}
\toprule
\textbf{Parameter} & \textbf{Value} \\
\midrule
Models & Qwen3.5-27B \\
Algorithm & GRPO with dense rewards \\
LoRA & Rank 16, $\alpha=32$, attention + MLP \\
Learning rate & $1 \times 10^{-4}$ \\
Training horizon & 4 epochs (primary arms); 330 steps (extended arms) \\
Group size ($G$) & 4 \\
Max turns & 30 per episode \\
Max tokens & 4{,}096 \\
KL coefficient & 0.001 \\
Checkpoints & Per epoch \\
Infrastructure & SageMaker p4d.24xlarge (8$\times$ A100 40GB) \\
\bottomrule
\end{tabular}
\end{table}

\paragraph{Rollout harness.} The harness operates as a rollout-as-a-service backend: (i) the training loop triggers a batch of $N$ tasks $\times$ group size episodes; (ii) episodes run in parallel Docker containers with LLM calls routed through a LiteLLM proxy that captures per-token logprobs to S3; (iii) hybrid verifiers produce rewards; (iv) the training loop fetches logprobs and rewards, runs a GRPO backward pass, and produces a LoRA update; (v) updated weights hot-swap to the SageMaker endpoint within $\sim$30s, no restart. Two model roles: a rollout model (Qwen3.5-27B on vLLM/SageMaker, generating completions with logprobs) and a frontier judge model used in the verification phase.

\section{Evaluation Setup}
\label{app:eval}

\paragraph{Held-out benchmarks.} Open-source harnesses pointed at trained-checkpoint serving endpoints: IFEval, MMLU-Pro, MATH500, GPQA $\to$ lm-eval-harness; BBEH $\to$ Inspect AI; HumanEval, DS-1000 $\to$ bigcode-eval-harness with sandboxed code execution; BFCL v4 and BFCL multi-turn $\to$ in-house rollout harness against trained-checkpoint endpoints; $\tau^2$-bench retail $\to$ Sierra's public \texttt{tau2-bench} simulator (\url{https://github.com/sierra-research/tau2-bench}) with the trained-checkpoint endpoint as the agent and the public user-model configuration. All ten benchmarks are evaluated on the trained model produced by each arm at the final training checkpoint, repeated across three decoding seeds per cell ($n=2$ for BFCL multi-turn).

\paragraph{Serving.} 3$\times$ g5.12xlarge SageMaker endpoints, TP=4, prefix caching enabled, 48 parallel decodes. Single-turn evals fit within standard inference budgets.

\paragraph{Decoding parameters.} Held-out benchmark generations use temperature $T=0.7$, top-$p$ $0.95$, top-$k$ $-1$, repetition penalty $1.0$, max new tokens $2{,}048$ (single-turn benchmarks) or $4{,}096$ (agentic multi-turn benchmarks), with three decoding seeds per cell. Frontier-LLM judge calls (reward verification, augmentation engine) use $T=1.0$, top-$p$ $0.95$, max tokens $2{,}048$, fixed model snapshot (Claude Opus 4.6) across all arms. Retries: 3, with exponential backoff on transient API errors; no retry on content-filter rejections.

\paragraph{Pass@$k$ estimator.} We use the unbiased pass@$k$ estimator from~\cite{chen2021codex} with 16 completions per task; pass@$k$ for $k \in \{1, 2, 4, 8, 16\}$ is computed by subsampling.

\section{Failure-mode Coverage Detail}
\label{app:failure_modes}

We tag every training rollout with an LLM-judge root-cause label drawn from the schema described in \S~\ref{app:verifier}. Raw counts per arm differ by an order of magnitude because rollout volume scales with arm task count. The comparison most relevant to ``does augmentation expose qualitatively different model behaviour?'' therefore requires equal-rollout normalization. Table~\ref{tab:failure_coverage} reports both the raw mode counts and a bootstrap-subsampled count at $n=472$ rollouts (matching H10\_A0, the smallest arm).

\begin{table}[ht]
\centering
\caption{Failure-mode coverage by arm. Raw counts are sensitive to rollout volume; the bootstrap-subsampled column reports mode counts after subsampling each arm to $n=472$ (mean over $1{,}000$ resamples). The model-side modes considered are \texttt{MODEL\_CORRECT}, \texttt{MODEL\_WRONG\_APPROACH}, \texttt{MODEL\_BUG}, \texttt{MODEL\_PARTIAL}; task-side modes such as \texttt{TASK\_PROMPT\_VERIFIER\_MISMATCH} are excluded from the model-side count.}
\label{tab:failure_coverage}
\small
\begin{tabular}{@{}lrccc@{}}
\toprule
\textbf{Arm} & \textbf{$n$ rollouts} & \textbf{raw modes ($\geq 3$ hits)} & \textbf{subsampled to $n=472$} & \textbf{unique at $n=472$} \\
\midrule
A (10 base) & 472 & 7 & 7.0 & 0 \\
B (97 human) & 3{,}008 & 10 & 7.4 & 0 \\
C (10+80 aug.) & 2{,}936 & 9 & 7.1 & 0--1 \\
D (10+319 aug.) & 5{,}456 & 12 & 7.6 & 1--2 \\
\bottomrule
\end{tabular}
\end{table}

After rollout-count normalization, the augmented arms still expose 0--2 additional model-side modes that the human-only arm does not, but the magnitude is small and within bootstrap variance for $n=472$ resamples. We include this analysis as a behavioural diagnostic; it does not, on its own, support a claim of ``broader capability coverage.''

\section{Quality-Gate Operating Decisions}
\label{app:decisions}

Pre-specified operating rules used during pipeline runs (independent of paper claims): if $<$250 variants pass $Q \geq 0.5$ in calibration, recalibrate the pipeline; at training step 100, if $>$40\% of H10\_A319 variants have $V_{\text{info}} < 0.1$, prune to the top 60\% by informativeness; if $V_{\text{verify}}$ stagnates while total reward climbs, reduce auxiliary reward weights.

\section{Compute Resources}
\label{app:compute}

All training arms ran on SageMaker p4d.24xlarge instances (8$\times$ A100 40GB) with a per-arm budget of approximately \$20K. The four primary arms and two compute-matched extended arms together consumed on the order of 900--1{,}200 GPU-hours of A100 time. Augmentation pipeline compute (variant generation + quality gating) used a mixture of frontier-LLM API calls. Evaluation runs on 3$\times$ g5.12xlarge SageMaker endpoints (TP=4, prefix-caching enabled, 48 parallel decodes). Total eval compute across 10 benchmarks $\times$ 6 trained checkpoints $\times$ 3 decoding seeds is on the order of \$300.

\section{Per-Arm Training-Internal Metrics}
\label{app:per_arm_metrics}

Per-arm reward + loss trajectories and per-arm full diagnostic panels (KL, entropy, gradient norm, within-group reward variance, GRPO degenerate-batch fraction, per-step time budget) are produced by a uniform analysis pipeline applied identically to each arm.

\begin{figure}[ht]
\centering
\includegraphics[width=\textwidth]{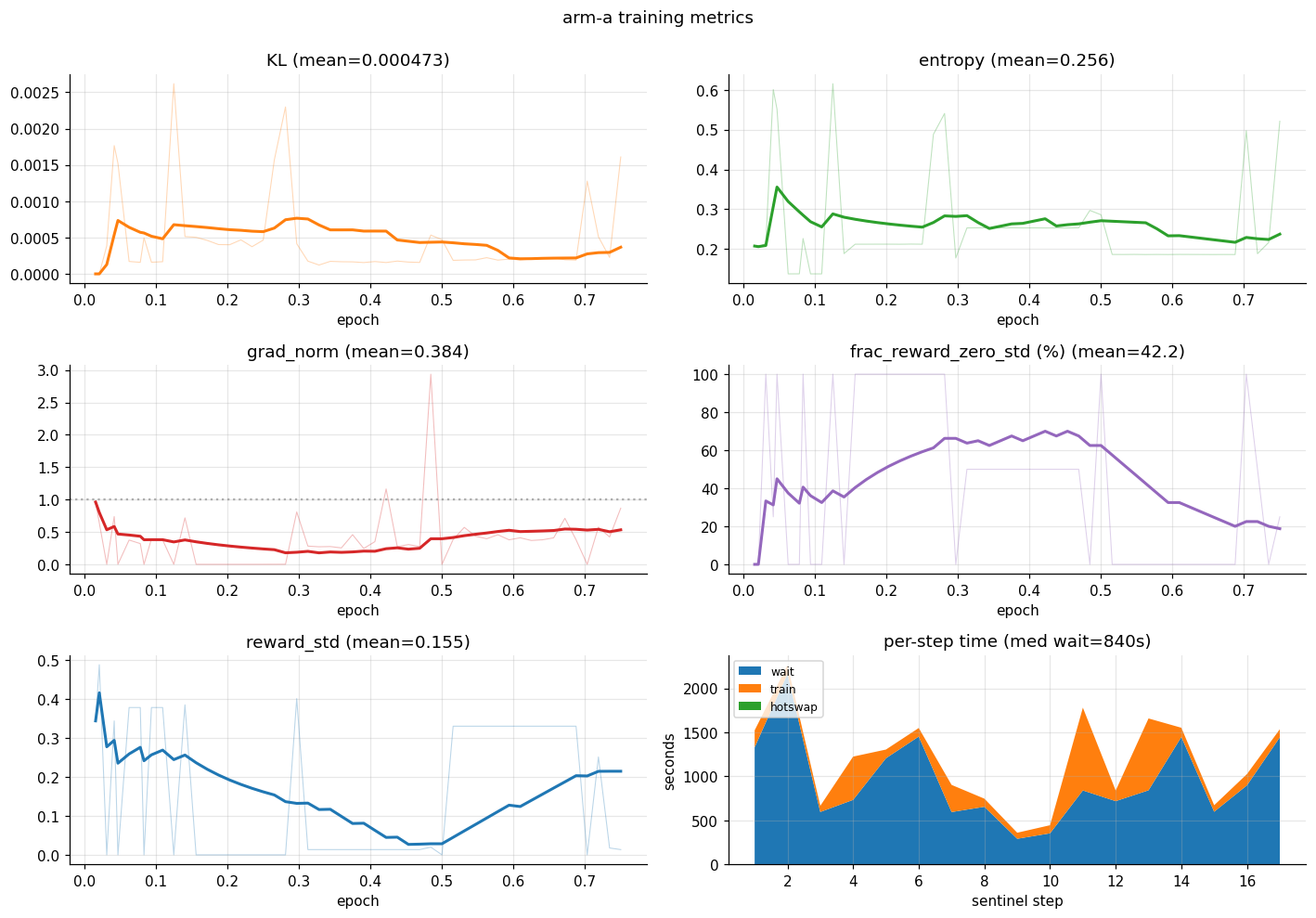}
\caption{H10\_A0 (10 base human tasks): training-internal metrics.}
\end{figure}

\begin{figure}[ht]
\centering
\includegraphics[width=\textwidth]{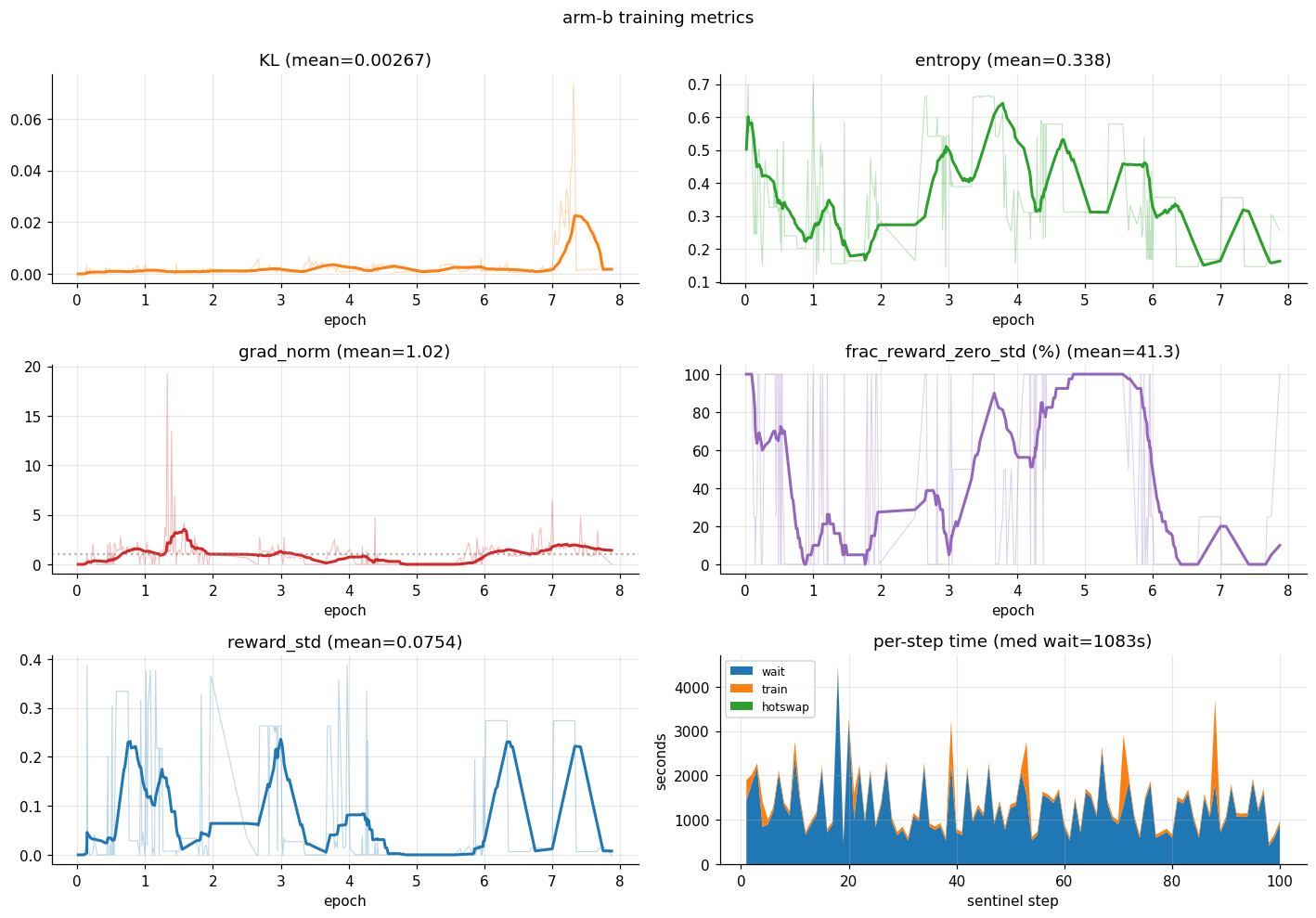}
\caption{H97\_A0 (97 human tasks): training-internal metrics.}
\end{figure}

\begin{figure}[ht]
\centering
\includegraphics[width=\textwidth]{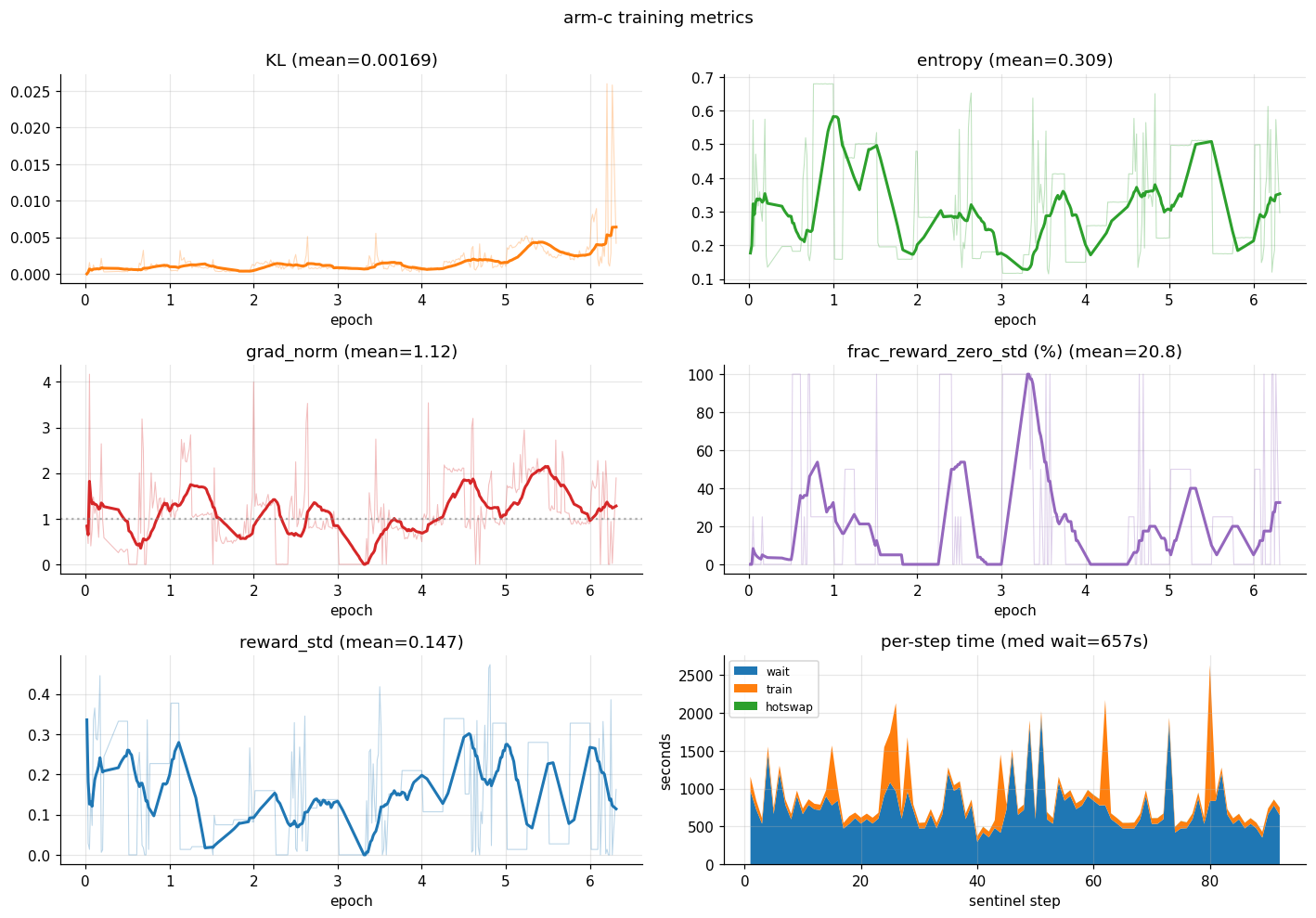}
\caption{H10\_A80 (10 base + 80 augmented, near-compute-equivalent to H97\_A0): training-internal metrics.}
\end{figure}

\begin{figure}[ht]
\centering
\includegraphics[width=\textwidth]{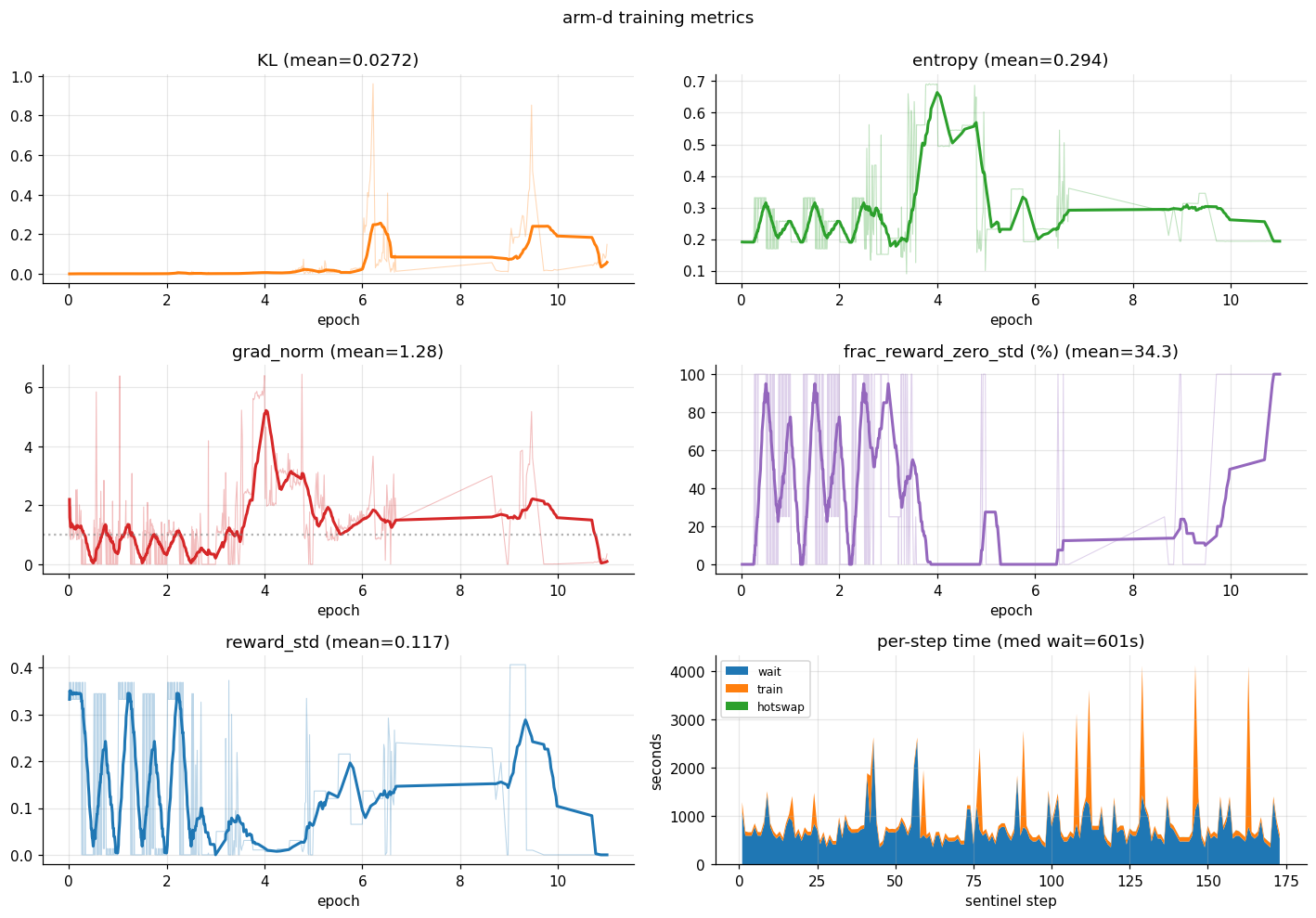}
\caption{H10\_A319 (10 base + 319 augmented; $4\times$ scaled augmentation over H10\_A80): training-internal metrics.}
\end{figure}

\section{Per-Arm Held-Out Benchmark Comparison}
\label{app:bench_paired}

Table~\ref{tab:bench_full} below reports the full per-benchmark pass@1 across all six arms (four primary + two compute-matched extended), as mean $\pm$ std across three decoding seeds per cell ($n=2$ for BFCL multi-turn). Table~\ref{tab:bench_untrained} records the untrained Qwen3.5-27B baseline per benchmark for reference; all four primary trained arms lift over this baseline on the ten-benchmark grand mean.

\begin{table*}[ht]
\centering
\caption{Full per-benchmark pass@1 (\%) across all six arms. Mean $\pm$ std over three decoding seeds ($n=2$ for BFCL multi-turn). H97\_A0$^{\dagger}$ and H10\_A80$^{\dagger}$ are trained to the H10\_A319 total step count under the same configuration. Body Table~\ref{tab:bench_main} reports the ten-benchmark grand-mean per arm; body Figures~\ref{fig:bench_arms} and~\ref{fig:dagger} visualize the primary near-step-matched (H10\_A80 vs.\ H97\_A0) and compute-matched (330-step) comparisons per benchmark.}
\label{tab:bench_full}
\footnotesize
\setlength{\tabcolsep}{2pt}
\begin{tabular}{@{}lcccccc@{}}
\toprule
\textbf{Bench} & \textbf{H10\_A0} & \textbf{H97\_A0} & \textbf{H10\_A80} & \textbf{H10\_A319} & \textbf{H97\_A0$^{\dagger}$} & \textbf{H10\_A80$^{\dagger}$} \\
\midrule
HumanEval~\cite{chen2021codex}         & $93.67 \pm 0.76$ & $94.70 \pm 0.35$ & $94.60 \pm 1.65$ & $94.73 \pm 1.99$ & $95.73 \pm 1.73$ & $94.51 \pm 0.86$ \\
DS-1000~\cite{ds1000}                  & $17.47 \pm 4.03$ & $16.73 \pm 5.02$ & $15.83 \pm 2.52$ & $18.07 \pm 3.98$ & $22.50 \pm 1.41$ & $18.75 \pm 2.47$ \\
IFEval~\cite{ifeval}                   & $85.00 \pm 0.00$ & $88.50 \pm 1.97$ & $89.70 \pm 0.17$ & $88.70 \pm 2.33$ & $88.48 \pm 0.84$ & $87.71 \pm 0.60$ \\
MATH500~\cite{math500}                 & $52.07 \pm 0.90$ & $50.20 \pm 3.62$ & $51.93 \pm 4.04$ & $52.47 \pm 2.83$ & $48.70 \pm 1.56$ & $45.70 \pm 0.14$ \\
BBEH~\cite{bbeh}                       & $ 1.83 \pm 0.06$ & $ 1.93 \pm 0.31$ & $ 1.90 \pm 0.17$ & $ 1.70 \pm 0.00$ & $ 1.80 \pm 0.28$ & $ 1.92 \pm 0.11$ \\
MMLU-Pro~\cite{mmlu_pro}               & $63.20 \pm 1.00$ & $63.00 \pm 1.81$ & $62.37 \pm 3.46$ & $63.80 \pm 2.50$ & $65.60 \pm 3.96$ & $64.10 \pm 1.27$ \\
GPQA Diamond~\cite{gpqa}               & $40.77 \pm 0.23$ & $41.53 \pm 0.99$ & $42.13 \pm 1.16$ & $41.93 \pm 1.93$ & $42.37 \pm 3.84$ & $43.06 \pm 2.14$ \\
BFCL v4~\cite{bfcl}                    & $59.44 \pm 0.00$ & $59.57 \pm 0.46$ & $59.17 \pm 0.84$ & $59.17 \pm 0.23$ & $59.04 \pm 0.00$ & $59.24 \pm 0.86$ \\
BFCL multi-turn~\cite{bfcl}            & $73.25 \pm 0.35$ & $71.25 \pm 0.35$ & $71.75 \pm 2.47$ & $73.25 \pm 1.77$ & $69.50 \pm 0.71$ & $72.50 \pm 0.71$ \\
$\tau^2$-bench retail~\cite{tau2bench} & $49.42 \pm 4.14$ & $54.39 \pm 0.88$ & $54.39 \pm 1.76$ & $57.60 \pm 1.83$ & $53.51 \pm 3.72$ & $53.56 \pm 3.10$ \\
\midrule
\textbf{Mean (10-bench grand-mean)} & $53.61 \pm 0.35$ & $54.18 \pm 0.40$ & $54.38 \pm 0.42$ & $\mathbf{55.14 \pm 0.41}$ & $54.72 \pm 0.42$ & $54.11 \pm 0.29$ \\
\bottomrule
\end{tabular}
\end{table*}

\begin{table}[ht]
\centering
\caption{Untrained Qwen3.5-27B baseline pass@1 (\%) per held-out benchmark. Reference for the lift-over-base computations in body Table~\ref{tab:bench_main}.}
\label{tab:bench_untrained}
\small
\setlength{\tabcolsep}{4pt}
\begin{tabular}{@{}lr@{}}
\toprule
\textbf{Bench} & \textbf{Untrained base} \\
\midrule
HumanEval~\cite{chen2021codex}     & 92.7 \\
DS-1000~\cite{ds1000}              & 13.5 \\
IFEval~\cite{ifeval}               & 88.8 \\
MATH500~\cite{math500}             & 52.2 \\
BBEH~\cite{bbeh}                   &  1.8 \\
MMLU-Pro~\cite{mmlu_pro}           & 62.7 \\
GPQA Diamond~\cite{gpqa}           & 43.2 \\
BFCL v4~\cite{bfcl}                & 59.0 \\
BFCL multi-turn~\cite{bfcl}        & 73.0 \\
$\tau^2$-bench retail~\cite{tau2bench} & 49.1 \\
\bottomrule
\end{tabular}
\end{table}

% Body Figure~\ref{fig:bench_arms} and Figure~\ref{fig:dagger} already provide
% the per-benchmark faceted breakdown for the primary and compute-matched
% comparisons; appendix versions removed to avoid duplication.

% Customer-facing memo: NeurIPS checklist suppressed; restore for venue submission.
% \clearpage
% \input{checklist.tex}


\begin{thebibliography}{99}

\bibitem{agent-rlvr}
Da, J., Wang, C., Deng, X., Ma, Y., Barhate, N., and Hendryx, S.
\newblock Agent-RLVR: Training software engineering agents via guidance and environment rewards.
\newblock \emph{arXiv:2506.11425}, 2025.
\newblock URL: \url{https://arxiv.org/abs/2506.11425}.

\bibitem{deepseek-r1}
DeepSeek-AI, Guo, D., Yang, D., Zhang, H., Song, J., et al.
\newblock DeepSeek-R1: Incentivizing reasoning capability in LLMs via reinforcement learning.
\newblock \emph{arXiv:2501.12948}, 2025.
\newblock URL: \url{https://arxiv.org/abs/2501.12948}.

\bibitem{med-rlvr}
Zhang, S., Liu, Q., Qin, G., Naumann, T., and Poon, H.
\newblock Med-RLVR: Emerging medical reasoning from a 3B base model via reinforcement learning.
\newblock \emph{arXiv:2502.19655}, 2025.
\newblock URL: \url{https://arxiv.org/abs/2502.19655}.

\bibitem{resyn}
He, A., Weir, N., Bostrom, K., Nie, A., Cassel, D., et al.
\newblock ReSyn: Autonomously scaling synthetic environments for reasoning models.
\newblock \emph{arXiv:2602.20117}, 2026.
\newblock URL: \url{https://arxiv.org/abs/2602.20117}.

\bibitem{agentrl}
Zhang, H., Liu, X., Lv, B., Sun, X., Jing, B., et al.
\newblock AgentRL: Scaling agentic reinforcement learning with a multi-turn, multi-task framework.
\newblock \emph{arXiv:2510.04206}, 2025.
\newblock URL: \url{https://arxiv.org/abs/2510.04206}.

\bibitem{prorl}
Zhang, H., Liu, M., Zhang, S., Han, S., Hu, J., et al.
\newblock ProRL Agent: Rollout-as-a-Service for RL training of multi-turn LLM agents.
\newblock \emph{arXiv:2603.18815}, 2026.
\newblock URL: \url{https://arxiv.org/abs/2603.18815}.

\bibitem{yue2025}
Yue, Y., Chen, Z., Lu, R., Zhao, A., Wang, Z., Song, S., and Huang, G.
\newblock Does reinforcement learning really incentivize reasoning capacity in LLMs beyond the base model?
\newblock \emph{arXiv:2504.13837}, 2025.
\newblock URL: \url{https://arxiv.org/abs/2504.13837}.

\bibitem{chen2021codex}
Chen, M., Tworek, J., Jun, H., Yuan, Q., Pinto, H. P. de O., et al.
\newblock Evaluating large language models trained on code.
\newblock \emph{arXiv:2107.03374}, 2021.
\newblock URL: \url{https://arxiv.org/abs/2107.03374}.

\bibitem{tinyv}
Xu, Z., Li, Y., Liu, Z., Yu, X., Wang, J., et al.
\newblock TinyV: Reducing false negatives in verification improves RL for LLM reasoning.
\newblock \emph{arXiv:2505.14625}, 2025.
\newblock URL: \url{https://arxiv.org/abs/2505.14625}.

\bibitem{ds1000}
Lai, Y., Li, C., Wang, Y., Zhang, T., Zhong, R., Zettlemoyer, L., Yih, S. W., Fried, D., Wang, S., and Yu, T.
\newblock DS-1000: A natural and reliable benchmark for data science code generation.
\newblock In \emph{Proceedings of the 40th International Conference on Machine Learning (ICML)}, 2023.
\newblock URL: \url{https://arxiv.org/abs/2211.11501}.

\bibitem{math500}
Lightman, H., Kosaraju, V., Burda, Y., Edwards, H., Baker, B., Lee, T., Leike, J., Schulman, J., Sutskever, I., and Cobbe, K.
\newblock Let's verify step by step.
\newblock In \emph{International Conference on Learning Representations (ICLR)}, 2024.
\newblock URL: \url{https://arxiv.org/abs/2305.20050}.

\bibitem{ifeval}
Zhou, J., Lu, T., Mishra, S., Brahma, S., Basu, S., Luan, Y., Zhou, D., and Hou, L.
\newblock Instruction-following evaluation for large language models.
\newblock \emph{arXiv:2311.07911}, 2023.
\newblock URL: \url{https://arxiv.org/abs/2311.07911}.

\bibitem{bbeh}
Kazemi, M., Fatemi, B., Bansal, H., Palowitch, J., Anastasiou, C., et al.
\newblock BIG-Bench Extra Hard.
\newblock In \emph{Proceedings of the 63rd Annual Meeting of the Association for Computational Linguistics (ACL)}, 2025.
\newblock \emph{arXiv:2502.19187}.
\newblock URL: \url{https://arxiv.org/abs/2502.19187}.

\bibitem{mmlu_pro}
Wang, Y., Ma, X., Zhang, G., Ni, Y., Chandra, A., et al.
\newblock MMLU-Pro: A more robust and challenging multi-task language understanding benchmark.
\newblock In \emph{Advances in Neural Information Processing Systems 37 (NeurIPS), Datasets and Benchmarks Track}, 2024.
\newblock URL: \url{https://arxiv.org/abs/2406.01574}.

\bibitem{gpqa}
Rein, D., Hou, B. L., Stickland, A. C., Petty, J., Pang, R. Y., Dirani, J., Michael, J., and Bowman, S. R.
\newblock GPQA: A graduate-level Google-proof Q\&A benchmark.
\newblock In \emph{Conference on Language Modeling (COLM)}, 2024.
\newblock URL: \url{https://arxiv.org/abs/2311.12022}.

\bibitem{bfcl}
Patil, S. G., Mao, H., Yan, F., Ji, C. C.-J., Suresh, V., Stoica, I., and Gonzalez, J. E.
\newblock The Berkeley Function-Calling Leaderboard (BFCL): From tool use to agentic evaluation of large language models.
\newblock In \emph{Proceedings of the 42nd International Conference on Machine Learning (ICML)}, PMLR 267:48371--48392, 2025.
\newblock URL: \url{https://proceedings.mlr.press/v267/patil25a.html}.

\bibitem{tau2bench}
Barres, V., Trinh, H., Yao, S., et al.
\newblock $\tau^2$-Bench: Evaluating conversational agents in a dual-control environment.
\newblock \emph{arXiv:2506.07982}, 2025.
\newblock URL: \url{https://arxiv.org/abs/2506.07982}.

\bibitem{embrace-error}
Krishna, R., Hata, K., Chen, S., Kravitz, J., Shamma, D. A., Fei-Fei, L., and Bernstein, M. S.
\newblock Embracing error to enable rapid crowdsourcing.
\newblock In \emph{Proceedings of the 2016 CHI Conference on Human Factors in Computing Systems}, pp.\ 3167--3179, 2016.
\newblock DOI: \href{https://doi.org/10.1145/2858036.2858115}{10.1145/2858036.2858115}.

\bibitem{hybrid-llm}
Ding, D., Mallick, A., Wang, C., Sim, R., Mukherjee, S., R\"uhle, V., Lakshmanan, L. V. S., and Awadallah, A. H.
\newblock Hybrid LLM: Cost-efficient and quality-aware query routing.
\newblock In \emph{International Conference on Learning Representations (ICLR)}, 2024.
\newblock \emph{arXiv:2404.14618}.
\newblock URL: \url{https://arxiv.org/abs/2404.14618}.

\bibitem{token-budget}
Han, T., Wang, Z., Fang, C., Zhao, S., Ma, S., and Chen, Z.
\newblock Token-budget-aware LLM reasoning.
\newblock In \emph{Findings of the Association for Computational Linguistics: ACL 2025}, 2025.
\newblock \emph{arXiv:2412.18547}.
\newblock URL: \url{https://aclanthology.org/2025.findings-acl.1274/}.

\bibitem{rad}
Laskin, M., Lee, K., Stooke, A., Pinto, L., Abbeel, P., and Srinivas, A.
\newblock Reinforcement learning with augmented data.
\newblock In \emph{Advances in Neural Information Processing Systems 33 (NeurIPS)}, 2020.
\newblock \emph{arXiv:2004.14990}.
\newblock URL: \url{https://arxiv.org/abs/2004.14990}.

\bibitem{drq}
Kostrikov, I., Yarats, D., and Fergus, R.
\newblock Image augmentation is all you need: Regularizing deep reinforcement learning from pixels.
\newblock In \emph{International Conference on Learning Representations (ICLR)}, 2021.
\newblock \emph{arXiv:2004.13649}.
\newblock URL: \url{https://arxiv.org/abs/2004.13649}.

\bibitem{paired}
Dennis, M., Jaques, N., Vinitsky, E., Bayen, A., Russell, S., Critch, A., and Levine, S.
\newblock Emergent complexity and zero-shot transfer via unsupervised environment design.
\newblock In \emph{Advances in Neural Information Processing Systems 33 (NeurIPS)}, 2020.
\newblock \emph{arXiv:2012.02096}.
\newblock URL: \url{https://arxiv.org/abs/2012.02096}.

\bibitem{plr}
Jiang, M., Grefenstette, E., and Rockt\"aschel, T.
\newblock Prioritized level replay.
\newblock In \emph{Proceedings of the 38th International Conference on Machine Learning (ICML)}, 2021.
\newblock \emph{arXiv:2010.03934}.
\newblock URL: \url{https://arxiv.org/abs/2010.03934}.

\bibitem{openassistant}
K\"{o}pf, A., Kilcher, Y., von R\"{u}tte, D., Anagnostidis, S., Tam, Z.-R., Stevens, K., Barhoum, A., Duc, N. M., Stanley, O., Nagyfi, R., et al.
\newblock OpenAssistant Conversations -- democratizing large language model alignment.
\newblock In \emph{Advances in Neural Information Processing Systems 36 (NeurIPS), Datasets and Benchmarks Track}, 2023.
\newblock URL: \url{https://arxiv.org/abs/2304.07327}.

\bibitem{tulu3}
Lambert, N., Morrison, J., Pyatkin, V., Huang, S., Ivison, H., Brahman, F., Miranda, L. J. V., Liu, A., Dziri, N., et al.
\newblock T\"{u}lu 3: Pushing frontiers in open language model post-training.
\newblock \emph{arXiv:2411.15124}, 2024.
\newblock URL: \url{https://arxiv.org/abs/2411.15124}.

\bibitem{swegym}
Pan, J., Wang, X., Neubig, G., Jaitly, N., Ji, H., Suhr, A., and Zhang, Y.
\newblock Training software engineering agents and verifiers with SWE-Gym.
\newblock In \emph{International Conference on Machine Learning (ICML)}, 2025.
\newblock URL: \url{https://arxiv.org/abs/2412.21139}.

\bibitem{self-instruct}
Wang, Y., Kordi, Y., Mishra, S., Liu, A., Smith, N. A., Khashabi, D., and Hajishirzi, H.
\newblock Self-Instruct: Aligning language models with self-generated instructions.
\newblock In \emph{Annual Meeting of the Association for Computational Linguistics (ACL)}, 2023.
\newblock URL: \url{https://arxiv.org/abs/2212.10560}.

\bibitem{evol-instruct}
Xu, C., Sun, Q., Zheng, K., Geng, X., Zhao, P., Feng, J., Tao, C., and Jiang, D.
\newblock WizardLM: Empowering large pre-trained language models to follow complex instructions.
\newblock In \emph{International Conference on Learning Representations (ICLR)}, 2024.
\newblock URL: \url{https://arxiv.org/abs/2304.12244}.

\bibitem{athey-imbens-2017}
Athey, S. and Imbens, G.~W.
\newblock The State of Applied Econometrics: Causality and Policy Evaluation.
\newblock \emph{Journal of Economic Perspectives}, 31(2):3--32, 2017.

\bibitem{saito-joachims-2022}
Saito, Y. and Joachims, T.
\newblock Counterfactual Evaluation and Learning for Interactive Systems.
\newblock Tutorial at the \emph{28th ACM SIGKDD Conference on Knowledge Discovery and Data Mining}, 2022.

\end{thebibliography}
\end{document}